\documentclass{article}

\PassOptionsToPackage{hyphens}{url}

\usepackage[utf8]{inputenc}
\usepackage{xcolor}
\usepackage{microtype}
\usepackage{graphicx}
\usepackage{subcaption}
\usepackage{booktabs}
\usepackage{hyperref}

\usepackage{amsmath}
\usepackage{amssymb}
\usepackage{amsthm}
\usepackage{ifthen}

\theoremstyle{plain}

\theoremstyle{definition}
\newtheorem{defn}{Definition}

\makeatletter
\newcommand*{\inlineequation}[2][]{%
  \begingroup
    \refstepcounter{equation}%
    \ifx\\#1\\%
    \else
      \label{#1}%
    \fi
    \relpenalty=10000 %
    \binoppenalty=10000 %
    \ensuremath{%
      #2%
    }%
    ~\@eqnnum
  \endgroup
}
\makeatother

\usepackage[accepted]{style/icml2019}

\usepackage{enumitem}
\setlist[itemize]{noitemsep,nolistsep}

\newcommand{\paperrunningtitle}{Distributed Momentum for Byzantine-resilient Learning}
\newcommand{\papertitle}{Distributed Momentum for Byzantine-resilient Learning}

\newcommand{\krumraw}{Krum}
\newcommand{\krum}{\textit{\krumraw{}}}
\newcommand{\mkrumraw}{Multi-Krum}
\newcommand{\mkrum}{\textit{\mkrumraw{}}}
\newcommand{\bulyanraw}{Bulyan}
\newcommand{\bulyan}{\textit{\bulyanraw{}}}
\newcommand{\bulyanrawof}[1]{\bulyanraw{} of {#1}}

\newcommand{\medianraw}{Median}
\newcommand{\medianoid}{\textit{\medianraw{}}}

\newcommand{\kardamraw}{Kardam}
\newcommand{\kardam}{\textit{\kardamraw{}}}

\newcommand{\byzgar}{\textit{F}}

\newcommand{\ldotexp}{\ldots{}}
\newcommand{\setr}{\mathbb{R}}
\newcommand{\setn}{\mathbb{N}}
\newcommand{\expect}{\mathop{{}\mathbb{E}}}

\newcommand{\card}[1]{\left\lvert{#1}\right\rvert}
\newcommand{\absv}[1]{\card{#1}}
\newcommand{\norm}[1]{\left\lVert{#1}\right\rVert}
\newcommand{\floor}[1]{\left\lfloor{#1}\right\rfloor}

\newcommand{\datadist}{\mathcal{D}}
\newcommand{\reallossletter}[1]{Q}
\newcommand{\datalossletter}[1]{q}
\newcommand{\realloss}[1]{\reallossletter{}\left({#1}\right)}
\newcommand{\dataloss}[2]{\datalossletter{}\left({#1}, {#2}\right)}

\newcommand{\indexed}[1]{\!\left[{#1}\right]}
\newcommand{\indexvar}[3]{{{#3}^{\ifthenelse{\equal{#1}{}}{}{\left({#1}\right)}}_{#2}}}

\newcommand{\paramsletter}{\theta}
\newcommand{\params}[1]{\indexvar{}{#1}{\paramsletter{}}}
\newcommand{\gradcomp}[2]{\indexvar{#1}{#2}{g}}
\newcommand{\gradsent}[2]{\indexvar{#1}{#2}{G}}
\newcommand{\graddist}[1]{\indexvar{}{#1}{\mathcal{G}}}
\newcommand{\gradnorm}[1]{\indexvar{}{#1}{\lambda}}
\newcommand{\gradsdev}[1]{\indexvar{}{#1}{\sigma}}
\newcommand{\mntmgradnorm}[1]{\indexvar{}{#1}{\Lambda}}
\newcommand{\mntmgradsdev}[1]{\indexvar{}{#1}{\Omega}}
\newcommand{\lr}[1]{\eta_{{#1}}}
\newcommand{\smallsim}{\ensuremath{\text{\raisebox{1pt}{$\scriptstyle \sim{}$}}}}

\DeclareMathOperator*{\argmin}{arg\,min}

\newcommand{\includeplot}[1]{\includegraphics[width=\linewidth,trim={6mm 6mm 6mm 6mm},clip]{plots/#1.jpg}}
\newcommand{\multicaption}[1]{\vspace{2mm}\caption{#1}\vspace{-2mm}}
\newcommand{\lrtitle}{$\lr{t} = \begin{cases}
    0.01  & \text{if }t < 1500 \\
    0.001 & \text{otherwise}
\end{cases}$}

\allowdisplaybreaks

\icmltitlerunning{\paperrunningtitle{}}

\begin{document}

\twocolumn[
\icmltitle{\papertitle{}}

\begin{icmlauthorlist}
\icmlauthor{El-Mahdi El-Mhamdi}{epfl}
\icmlauthor{Rachid Guerraoui}{epfl}
\icmlauthor{S\'{e}bastien Rouault}{epfl}
\end{icmlauthorlist}

\icmlaffiliation{epfl}{Distributed Computing Laboratory (DCL), \'{E}cole Polytechnique F\'{e}d\'{e}rale de Lausanne (EPFL), Switzerland}

\icmlcorrespondingauthor{S\'{e}bastien Rouault}{sebastien.rouault@epfl.ch}

\icmlkeywords{Machine Learning, Distributed, Byzantine resilience, Momentum}

\vskip 0.3in
]

\printAffiliationsAndNotice{Author list in alphabetical order.}

\begin{abstract}
Momentum is a variant of gradient descent that has been proposed for its benefits on convergence.
In a distributed setting, momentum can be implemented either at the server or the worker side.
When the aggregation rule used by the server is linear, commutativity with addition makes both deployments equivalent.
Robustness and privacy are however among motivations to abandon linear aggregation rules.
In this work, we demonstrate the benefits on robustness of using momentum at the worker side.
We first prove that computing momentum at the workers reduces the variance-norm ratio of the gradient estimation at the server, strengthening Byzantine resilient aggregation rules.
We then provide an extensive experimental demonstration of the robustness effect of worker-side momentum on distributed SGD.

\end{abstract}

\setlength\abovedisplayskip{\parskip}
\setlength\belowdisplayskip{\parskip}

\section{Introduction}
\label{sec:introduction}








Gradient descent is the driving force of the recent successes in machine learning.
Large-scale deployment of gradient descent relies on two ideas: stochastic approximation and distribution.
Stochastic approximation (drastically) reduces the computation time, at the price of introducing variance in the gradient estimations.
Distribution alleviates the workload on a single machine but, as we discuss below, the multiplicity of elements inevitably increases the likelihood of (malicious) faults.

\begin{figure}
    \centering
    \includegraphics[width=\linewidth]{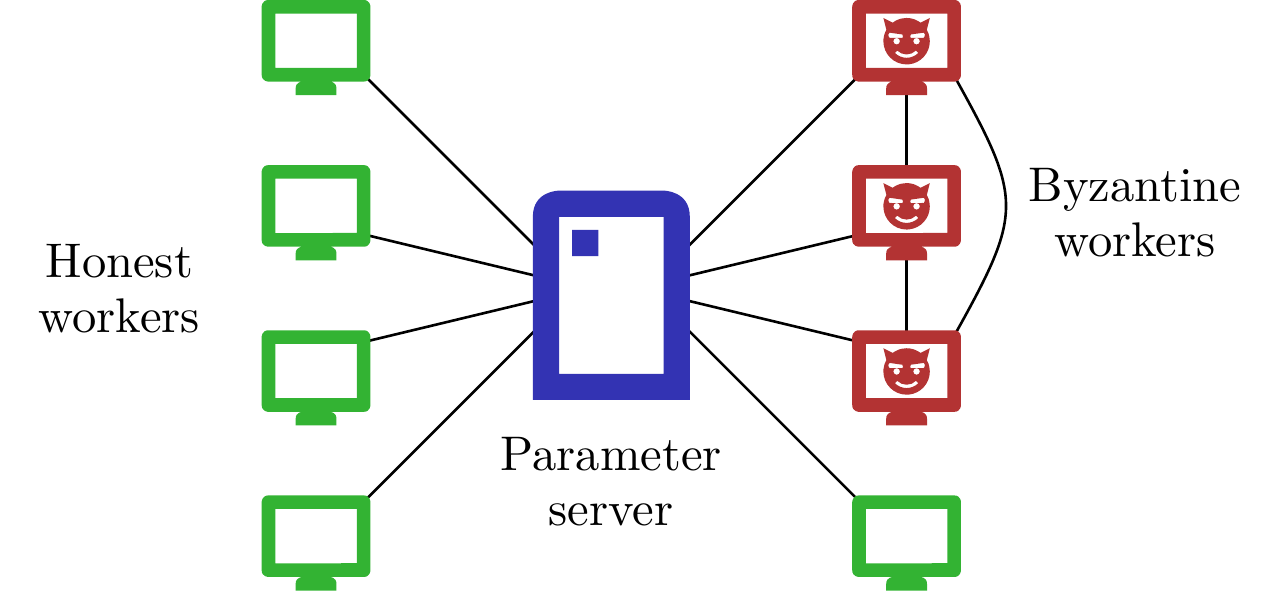}
    \caption{A parameter server setup with $n = 8$ workers, among which $f = 3$ are \emph{Byzantine} (i.e., adversarial) workers.
    A black line represents a bidirectional communication channel.}
    \label{fig:background-model}
\end{figure}



In the distributed parameter server setting, the training of a model is basically performed as follows.
A central machine, called the {\it server}, sends the current model (the vector of parameters) to other machines, called {\it workers}.
These use their share of data (either their local and private data, or data provided by the server for training purpose) to compute a gradient estimate which is in turn sent to the server.
As the server receives the gradients from different workers, the server typically averages their values to update the model if the setting is synchronous, or updates the model as individual gradients are received if the model is asynchronous.

When all the workers are reliable and provide correct estimates of the gradient, this setting has close to optimal behavior~\cite{lian2015asynchronous, zhang2016asynchronous, dean2012large}.
Many practical factors could however make the correctness assumption of the workers doubtful.
These factors span a large spectrum of causes, from software bugs, noisy or poisonous data, stale machines or worse, malicious attackers controlling some machines.

The Byzantine abstraction is a very general fault model in distributed systems~\cite{byzantine-lamport}. 
The standard, golden solution for Byzantine fault tolerance is the \emph{state machine replication} approach~\cite{SMR}.
This approach is however based on \emph{replication}, which is~unsuitable for distributed machine learning and stochastic gradient descent, such as
%
federated learning~\cite{federated-learning}.
Workers could be independent entities, who could not be replicated for obvious privacy, scalability or legal reasons.
For instance, in the context of federated learning, recent work has shown that Byzantine fault tolerance serves as a good basis to study poisoning~\cite{federated-backdoor, federated-no-backdoor}. In that same context, recent results show that Byzantine-resilient aggregation rules are effective against {\it distributed backdoor attacks}~\cite{xie2020dba}.




The key vulnerability in the standard parameter server rests upon how gradients are aggregated.
Since 2017, many alternatives to averaging have been proposed: \cite{byzantine-alistarh, draco, median-optimal, median, krum, bulyan, kardam, yang2019byrdie, tianxiang2019aggregation, signsgd, median, yang2019bridge, draco, xie2018phocas, yang2019adversary, detox, munoz2019byzantine} to list a few.
In synchronous settings, these solutions 
consists in replacing the averaging of gradients by a robust alternative such as the median and its variants~\cite{krum, bulyan, median} or~redundancy schemes~\cite{draco, detox}. In asynchronous settings, since no aggregation can be made, gradients are (ideally) used individually as they are delivered, the robust alternatives are less diverse and are mostly consisting of a filtering scheme~\cite{kardam}.



One common aspect underlying these methods is their reliance on ``quality gradients'' from the non-Byzantine workers.
Technically: the variance between non-Byzantine
gradient estimates must be bounded below a factor of their average norm.
This requirement is not new in machine learning \cite{bottou}, and is actually independent from Byzantine considerations as an unbounded \emph{variance-norm ratio} would prevent convergence.


Is there a way to guarantee ``quality gradient'' at the non Byzantine workers?
Addressing this question is crucial to put Byzantine-resilient gradient descent to work.









We provide a positive answer to this question by using momentum~\cite{momentum}.
Momentum consists in summing a series of past gradients with the new one using an exponential decay factor $\mu$ ($0 < \mu < 1$), instead of using the new gradient alone.
Momentum can be computed at the server side, when the update is performed, or at the workers' side, when gradients are still computed~\cite{momentumdistribuee}. In non Byzantine-resilient settings, both deployments are equivalent, as the gradient aggregation used at the server is linear and commutes with addition.
In practice, momentum is typically employed at the server side.
In this work, we propose to use momentum at the workers' side since none of the existing Byzantine-resilient aggregation rules is linear.

We first show theoretically that indeed we can guarantee ``quality gradient'' by using momentum at the workers.
Then we report on an extensive experimental assessment of this claim.
In particular, and while using momentum at the workers has \emph{no additional overhead} over momentum at the server, this technique led to an observed $\times{}11$ reduction on cross-accuracy drop due to Byzantine actors~(Section~\ref{sec:experiments-results}).










\paragraph{Contributions.}
Essentially, we show for the first time that applying momentum at the workers significantly boosts robustness against
Byzantine behavior.
We prove that computing momentum at the workers reduces the \emph{variance-norm} ratio of the honest gradient estimations at the server, a key quantity for any robust alternative to averaging which approximates a high-dimensional median; for instance \cite{krum, median, bulyan}.
In particular, we show that combining now-standard defense mechanisms \cite{krum, median, bulyan} with momentum
(at the worker side) ensures previously unavailable safety guarantees and counters state-of-the-art attacks such as \cite{little, empire}.
We report on an extensive experimental evaluation of this claim with 88 different tested sets of hyperparameters (440 trained models in total), spanning the 2 mentioned state-of-the-art attacks and 3 defenses.


\paragraph{Paper Organization.}

Section~\ref{sec:background} formalizes the problem and provides the necessary background.
Section~\ref{sec:momentum} presents our theoretical contribution and compares
the usage of momentum at the workers versus at the server.
Section~\ref{sec:experiments} describes our experimental settings in details,
before presenting and analysing our experimental results.
Section~\ref{sec:conclusion} discusses related and future work.

Due to space limitation, only a representative fraction of the experimental results is presented in the main paper.
The supplementary material 
reports on the entirety of our experiments, along with the code and procedure to reproduce all of our results (including the graphs).

\section{Background}
\label{sec:background}

\subsection{Byzantine Distributed SGD}
\label{sec:background-bdsgd}

\paragraph{Stochastic Gradient Descent (SGD).}
We consider the classical problem of optimizing a non-convex, differentiable loss function $\reallossletter{}: \setr{}^d \rightarrow{} \setr{}$,
where $\realloss{\params{t}} \triangleq \expect{}_{x \sim{} \datadist{}}\left[ \dataloss{\params{t}}{x} \right]$ for a fixed data distribution $\datadist{}$.
Namely, we seek a $\paramsletter{}^{*} \in \setr{}^d$ such that:
\inlineequation[eq:background-convergence]{\nabla{}\realloss{\paramsletter{}^{*}} = 0}

Using SGD, we initially pick a random $\params{0} \in \setr{}^d$.
Then at every step $t \ge 0$, we uniformly sample $b$ datapoints $x_1 \ldots{} x_b$ from $\datadist{}$ to estimate a \emph{gradient} $\gradcomp{}{t} \triangleq \frac{1}{b}\sum_{k = 1}^{b}{\nabla{}\dataloss{\params{t}}{x_k}} \approx \nabla{}\realloss{\params{t}}$.
Finally, for step $t$, we update the parameter vector using $\params{t + 1} = \params{t} - \lr{t} \gradcomp{}{t}$, where $\lr{t} > 0$ is the \emph{learning rate}.

One field-tested amendment to this update rule is \emph{momentum} \cite{momentum},
where each gradient has an exponentially-decreasing effect on every subsequent update.
Formally: $\params{t + 1} = \params{t} - \lr{t} \sum_{u = 0}^{t}{\mu^{t - u} \gradcomp{}{u}}$,
with $0 < \mu < 1$.

\paragraph{Distributed SGD with Byzantine workers.}
We follow the parameter server model \cite{parameter-server}: $1$ process (the \emph{parameter server}) holding the parameter vector $\params{t} \in \setr{}^d$, and $n$ other processes (the \emph{workers}) estimating gradients.
Among these $n$ workers, up to $f < n$ are said \emph{Byzantine}, i.e., adversarial.
Unlike the other $n - f$ \emph{honest} workers, these $f$ \emph{Byzantine} workers can send arbitrary gradients (Figure \ref{fig:background-model}).


At each step $t$, the parameter server receives $n$ different gradients $\gradcomp{1}{t} \ldots{} \gradcomp{n}{t}$, among which $f$ are arbitrary (sent by the Byzantine workers).
So the update equation becomes:
\begin{flalign}
    && \params{t + 1} &= \params{t} - \lr{t} \gradsent{}{t} & \nonumber \\
    \text{where:} &&
    \gradsent{}{t} &\triangleq \sum\limits_{u=0}^{t}{\mu^{t - u} \byzgar\left( \gradcomp{1}{u}, \ldots{}, \gradcomp{n}{u} \right)} & \label{eq:background-at-update} \\
    \text{and where:} && \byzgar{}&: \left(\setr{}^d\right)^n \rightarrow{} \setr{}^d & \nonumber
\end{flalign}

The function $\byzgar{}$ is called a \emph{Gradient Aggregation Rule} (GAR).
If we assume no Byzantine worker, averaging is sufficient;
formally: $\byzgar{}\left( \gradcomp{1}{t}, \ldots{}, \gradcomp{n}{t} \right) = \frac{1}{n} \sum_{i = 1}^{n}{\gradcomp{i}{t}}$.
In the presence of Byzantine workers, a more complex aggregation is performed with a \emph{Byzantine-resilient} GAR.
Section \ref{sec:background-gar} presents the three Byzantine-resilient GARs studied in this paper, along with their own theoretical requirements.

\paragraph{Adversarial Model.}
The goal of the adversary is to impede the learning process, which can generally be defined as the maximization of the loss $\reallossletter{}$ or, more judiciously in the image classification tasks used in this paper, as the minimization\footnote{I.e., with $10$ classes, the worst possible final accuracy is $0.1$.} of the model's top-1 cross-accuracy.

The adversary cannot directly overwrite $\params{t}$ at the parameter server.
The adversary only submits $f$ arbitrary gradients to the server per step, via the $f$ Byzantine workers it controls\footnote{Said otherwise, the $f$ Byzantine workers can collude.}.

We assume an omniscient adversary.
In particular, the adversary knows the GAR used by the parameter server and can generate Byzantine gradients dependent on the honest gradients submitted at the same step.

\subsection{Byzantine-resilient GARs}
\label{sec:background-gar}

We formally present below the $3$ GARs studied in this paper.

These GARs are \emph{Byzantine-resilience}, a notion first introduced by \cite{krum} under the name $\left( \alpha, f \right)$-Byzantine-resilience.
When used within its operating assumptions, a Byzantine-resilient GAR guarantees convergence (in the sense of \eqref{eq:background-convergence}) even in an adversarial setting.

\begin{defn} \label{def:background-byz-res}
Let $\left( \alpha, f \right) \in \left[ 0 .. \frac{\pi}{2} \right[ \times \left[ 0 .. n \right]$, with $n$ the total number of workers.
Let $\left( \gradcomp{1}{t} \ldots{} \gradcomp{n}{t} \right) \in \left( \setr{}^d \right)^{n}$, among which $n - f$ are independent (``honest'') vectors following the same distribution $\graddist{t}$; the $f$ other vectors are arbitrary, each possibly dependent on $\graddist{t}$ and the ``honest'' vectors.

A GAR $\byzgar{}$ is said to be $\left( \alpha, f \right)$-Byzantine resilient iff:
\begingroup
\setlength\abovedisplayskip{\parskip}
\setlength\belowdisplayskip{\parskip}
\begin{equation*}
    \gradcomp{}{t} \triangleq \byzgar{}\left( \gradcomp{1}{t}, \ldots{}, \gradcomp{n}{t} \right)
\end{equation*}
satisfies:
\begin{enumerate}[nolistsep,noitemsep]
    \item{$\left\langle \expect\gradcomp{}{t}, \expect\graddist{t} \right\rangle \ge \left( 1 - \sin{\alpha} \right) \cdot \norm{\expect\graddist{t}} > 0$}
    \item{$\forall r \in \left\lbrace 2, 3, 4 \right\rbrace$, $\expect\norm{\gradcomp{}{t}}^r$ is bounded above by a linear combination of the terms $\expect\norm{\graddist{t}}^{r_1} \ldots{} \expect\norm{\graddist{t}}^{r_k}$, with $\left( k, r_1 \ldots{} r_k \right) \in \left( \setn{}^{*} \right)^{k + 1}$ and $r_1 + \ldots{} + r_k = r$.}
\end{enumerate}
\endgroup
\end{defn}

\subsubsection{\krumraw{} \cite{krum}}
\label{sec:background-gar-krum}

Let $\left( f, m \right) \in \setn{}^2$, with $n \ge 2 f + 3$ and $1 \le m \le n - f - 2$.

\krum{} works by assigning a score to each input gradient.
The score of $\gradcomp{i}{t}$ is the sum of the distances between $\gradcomp{i}{t}$ and its $n - f - 2$ closest neighbor gradients.
\krum{} outputs the arithmetic mean of the $m$ smallest--scoring gradients\footnote{The original paper called the GAR \mkrum{} when $m > 1$.}.

In our experiments, we set $m$ to its maximum: $n - f - 2$.

To be proven $\left( \alpha, f \right)$-Byzantine resilient, besides the standard convergence conditions in non-convex optimization \cite{bottou},
\krum{} requires the honest gradients' variance $\expect\norm{\graddist{t} - \expect\graddist{t}}^2$ to be bounded above as follows:
\begingroup
\setlength\abovedisplayskip{\parskip}
\setlength\belowdisplayskip{\parskip}
\begin{align}
    && & 2 \cdot \kappa\!\left( n, f \right) \cdot \expect\norm{\graddist{t} - \expect\graddist{t}}^2 < \norm{\expect\graddist{t}}^2\hspace*{-1cm} & \label{eq:background-ratio-bound} \\
    \text{with:} && & \kappa\!\left( n, f \right) \triangleq n\!-\!f\!+\!\frac{f \left( n\!-\!f\!-\!2 \right)\!+\!f^2 \left( n\!-\!f\!-\!1 \right)}{n\!-\!2 f\!-\!2} & \nonumber
\end{align}
\endgroup

\subsubsection{\medianraw{} \cite{median}}
\label{sec:background-gar-median}

\begingroup
\setlength\abovedisplayskip{\parskip}
\setlength\belowdisplayskip{\parskip}

Let $f \in \setn{}$ with $n \ge 2 f + 1$.

\medianoid{} computes the coordinate-wise median of the input gradients $\gradcomp{1}{t} \ldots{} \gradcomp{n}{t}$.
Formally for the real-valued median:
\begin{equation*}
    \text{median}\left( x_1 \ldotexp x_n \right) \triangleq \argmin_{x \in \setr{}}{\sum\limits_{i = 1}^{n}{\absv{x_i - x}}}
\end{equation*}

And so, formally for the coordinate--wise \medianoid{}:
\begin{equation*}
    \text{\medianraw{}} \left( \gradcomp{1}{t} \ldotexp \gradcomp{n}{t} \right) \triangleq \left( \substack{
    \text{median}\left( \gradcomp{1}{t}\indexed{1} \ldotexp \gradcomp{n}{t}\indexed{1} \right) \\[-1mm]
    \vdots \\[1mm]
    \text{median}\left( \gradcomp{1}{t}\indexed{d} \ldotexp \gradcomp{n}{t}\indexed{d} \right) } \right)
\end{equation*}

\endgroup

The condition of $\left( \alpha, f \right)$-Byzantine resilience is:
\begingroup
\setlength\abovedisplayskip{\parskip}
\setlength\belowdisplayskip{\parskip}
\begin{equation} \label{eq:background-ratio-bound-median}
    \left( n - f \right) \cdot \expect\norm{\graddist{t} - \expect\graddist{t}}^2 < \norm{\expect\graddist{t}}^2
\end{equation}
\endgroup

\subsubsection{\bulyanraw{} \cite{bulyan}}
\label{sec:background-gar-bulyan}

\bulyanraw{} uses another Byzantine-resilient GAR to aggregate the input gradients.
In the remaining of this paper we will consider \bulyanrawof{\krum{}}, that we will simply call \bulyan{}.

Let $\left( f, m \right) \in \setn{}^2$, with $n \ge 4 f + 3$ and $1 \le m \le n - f - 2$.

\bulyan{} first selects $n - 2 f - 2$ gradients by iterating $n - 2 f - 2$ times over \krum{}, each time removing the highest scoring gradient from the input gradient set.
From these $n - 2 f - 2 \ge 2 f + 1$ selected gradients, \bulyan{} outputs the coordinate-wise average of the $n - 4 f - 2 \ge 1$ closest coordinate values to the coordinate-wise median.

The theoretical requirements for the $\left( \alpha, f \right)$-Byzantine resilience of \bulyan{} are the same as the ones of \krum{}.

\subsection{Studied Attacks}
\label{sec:background-attack}

The two, state-of-the-art attacks studied in this paper follow the same core algorithm,
that we identify below. 

Let $\varepsilon_t \in \setr{}_{\ge 0}$ be a non-negative factor, and $a_t \in \setr{}^d$ an \emph{attack vector} which value depends on the actual attack used.

At each step $t$, each of the $f$ Byzantine workers submits the same Byzantine gradient: \inlineequation[eq:background-attack]{\overline{\gradcomp{}{t}} + \varepsilon_t \cdot a_t},
where $\overline{\gradcomp{}{t}}$ is an approximation of the real gradient $\nabla{}\realloss{\params{t}}$ at step $t$.

For both of the studied attacks, the value of $\varepsilon_t$ is fixed.

\subsubsection{A Little is Enough \cite{little}}
\label{sec:background-attack-little}

In this attack, a Byzantine worker submits $\overline{\gradcomp{}{t}} + \varepsilon_t \cdot a_t$, with $a_t \triangleq -\sigma_t$ the opposite of the coordinate-wise standard deviation of the honest gradient distribution $\graddist{t}$.


\subsubsection{Fall of Empires \cite{empire}}
\label{sec:background-attack-empire}

A Byzantine worker submits $\left( 1 - \varepsilon_t \right) \overline{\gradcomp{}{t}}$, i.e., $a_t \triangleq -\overline{\gradcomp{}{t}}$.

\section{Momentum at the Workers}
\label{sec:momentum}

The Byzantine-resilience of \krum{}, \medianoid{} and \bulyan{} rely on the honest gradients being sufficiently \emph{clumped}.
For the GARs we study, this is formalized in equations \eqref{eq:background-ratio-bound} and \eqref{eq:background-ratio-bound-median}.


This requirement is theoretically important.
When the variance of the honest gradients is \emph{too high} compared to their norms (e.g., Equation \eqref{eq:background-ratio-bound} unsatisfied), the Byzantine gradients can induce aggregated gradients having negative dot-products with the real gradient, preventing convergence (as such aggregated gradients would locally increase the loss).

This requirement is also not satisfied in practice.
When reproducing the attacks (figures \ref{fig:experiments-first} to \ref{fig:experiments-last}), we measured and observed that the honest gradients' variance is often at least one order of magnitude too large for all the studied GARs.
In the 400 experiments we performed under attack, we measured the theoretical requirement of Equation \eqref{eq:background-ratio-bound} is never satisfied, not even for a single step, in 394 of them.
Among the 6 other experiments (all with the CIFAR-10 model), equations \eqref{eq:background-ratio-bound} or \eqref{eq:background-ratio-bound-median} were never verified for more than 4 steps (out of 3000) per experiment.

Nevertheless, our empirical evidences show that \emph{reducing} the honest gradients' variance relative to their norm can be enough to defend the training against the two presented attacks.
In this section we present a technique aiming at decreasing the \emph{variance-norm ratio} of the honest gradients, reducing or even cancelling at negligible computational costs the effects of the attacks studied in this paper.

\subsection{Formulation}

From the formulation of \emph{momentum SGD} (Equation \eqref{eq:background-at-update}):
\begin{equation*}
    \gradsent{}{t} \triangleq \sum\limits_{u=0}^{t}{\mu^{t - u} \byzgar\left( \gradcomp{1}{u}, \ldots{}, \gradcomp{n}{u} \right)}
\end{equation*}
we instead confer the momentum operation on the workers:
\begin{equation} \label{eq:momentum-at-workers}
    \gradsent{}{t} \triangleq \byzgar\biggl( \underbrace{\sum\limits_{u=0}^{t}{\mu^{t - u} \gradcomp{1}{u}}}_{\gradsent{1}{t}}, \ldots{}, \underbrace{\sum\limits_{u=0}^{t}{\mu^{t - u} \gradcomp{n}{u}}}_{\gradsent{n}{t}} \biggr)
\end{equation}

\paragraph{Notations.}
In the remaining of this paper, we call the original formulation \emph{(momentum) at the server}, and the proposed, revised formulation \emph{(momentum) at the workers}.

\subsection{Effects}
\label{sec:momentum-effects}



We compare the variance-norm ratio when momentum is computed \emph{at the server} versus \emph{at the workers}.

Let $\gradnorm{t} \triangleq \norm{\expect\graddist{t}} > 0$ be the real gradient's norm at step $t$.\\%
Let $\gradsdev{t} \triangleq \sqrt{\expect\norm{\graddist{t} - \expect\graddist{t}}^2}$ be the standard deviation of the real gradient at step $t$.
The variance-norm ratio, when momentum is computed \emph{at the server}, is:
\begin{equation*}\label{eq:momentum-ratio-update}
    r_t^{(s)} \triangleq \frac{\gradsdev{t}^2}{\gradnorm{t}^2}
\end{equation*}

We will now compute this ratio when momentum is applied at the workers.
Let $\gradsent{i}{t}$, with $\gradsent{i}{-1} \triangleq 0$, be the gradient sent by any honest worker $i$ at step $t$, i.e.:
\begin{equation*}
    \gradsent{i}{t} \triangleq \sum\limits_{u=0}^{t}{\mu^{t - u} \gradcomp{i}{u}}
\end{equation*}
Then, for any two honest worker identifiers $i \neq j$:
\begin{align}
    &\phantom{=\null} \expect\norm{\gradsent{i}{t} - \gradsent{j}{t}}^2 \nonumber\\
    &= \expect\norm{\gradcomp{i}{t} + \mu \, \gradsent{i}{t - 1} - \gradcomp{j}{t} - \mu \, \gradsent{j}{t - 1}}^2 \nonumber\\
    &= \expect\norm{\gradcomp{i}{t} - \gradcomp{j}{t}}^2 + \mu^2 \, \expect\norm{\gradsent{i}{t - 1} - \gradsent{j}{t - 1}}^2 \nonumber\\
    &\phantom{=} + \underbrace{2 \, \mu \left( \underbrace{\expect\gradcomp{i}{t} - \expect\gradcomp{j}{t}}_{=\,\expect\graddist{t} - \expect\graddist{t}} \right) \cdot \left( \expect\gradsent{i}{t - 1} - \expect\gradsent{j}{t - 1} \right)}_{=\,0} \nonumber\\
    &= \expect\norm{\gradcomp{i}{t} - \gradcomp{j}{t}}^2 + \mu^2 \, \expect\norm{\gradsent{i}{t - 1} - \gradsent{j}{t - 1}}^2 \nonumber\\
    &= 2 \, \gradsdev{t}^2 + \mu^2 \left( 2 \, \gradsdev{t-1}^2 + \mu^2 \left( 2 \, \gradsdev{t-2}^2 + \mu^2 \left( ... \right) \right) \right) \nonumber\\
    &= 2 \, \sum_{u = 0}^{t}{\mu^{2 \left( t - u \right)} \gradsdev{u}^2} \label{eq:momentum-variance} \\
    &= 2 \, \expect\norm{\gradsent{i}{t} - \expect\gradsent{i}{t}}^2 \nonumber
\end{align}
\begin{align*}
    &\phantom{=\null} \norm{\expect\gradsent{i}{t}}^2
    = \norm{\expect\gradcomp{i}{t} + \mu \, \expect\gradsent{i}{t - 1}}^2 \\
    &= \norm{\expect\gradcomp{i}{t}}^2 + 2 \, \mu \, \expect\gradcomp{i}{t} \cdot \expect\gradsent{i}{t - 1} + \mu^2 \, \norm{\expect\gradsent{i}{t - 1}}^2 \\
    &= \gradnorm{t}^2 + 2 \, \mu \, \expect\gradcomp{i}{t} \cdot \left( \expect\gradcomp{i}{t - 1} + \mu \, \left( \expect\gradcomp{i}{t - 2} + \mu \, \left( ... \right) \right) \right) \\
    &\phantom{=} + \mu^2 \, \left( \gradnorm{t - 1}^2 + 2 \, \mu \, \expect\gradcomp{i}{t - 1} \cdot \left( \expect\gradcomp{i}{t - 2} + \mu \, \left( ... \right) \right) \right. \\
    &\phantom{\phantom{=} + \mu^2 \, \left(\right.} \left. + \mu^2 \, \expect\norm{\gradsent{i}{t - 2}}^2 \right) \\
    &= \sum_{u = 0}^{t}{\mu^{2 \left( t - u \right)} \left( \gradnorm{u}^2 + 2 \, \sum_{v = 0}^{u - 1}{\mu^{u - v} \underbrace{\expect\gradcomp{i}{u} \cdot \expect\gradcomp{i}{v}}_{= \expect\graddist{u} \cdot \expect\graddist{v}}} \right)}
\end{align*}

Thus, assuming honest gradients $\expect\gradsent{i}{t}$ do not become null:
\begin{equation*}
    r_t^{(w)} \triangleq \frac{\mntmgradsdev{t}^2}{\mntmgradnorm{t}^2} = \frac{\sum_{u = 0}^{t}{\mu^{2 \left( t - u \right)} \gradsdev{u}^2}}
    {\sum_{u = 0}^{t}{\mu^{2 \left( t - u \right)} \left( \gradnorm{u}^2 + s_u \right)}}
\end{equation*}
where the expected ``straightness'' of the gradient computed by an honest worker at step $u$ is defined by:
\begin{equation*}
    s_u \triangleq{} 2 \, \sum_{v = 0}^{u - 1}{\mu^{u - v} \expect\graddist{u} \cdot \expect\graddist{v}}
\end{equation*}

$s_u$ quantifies what can be thought as the \emph{curvature} of the honest gradient trajectory.
Straight trajectories can make $s_u$ grow up to $\left( 1 - \mu \right)^{-1}\!>\!1$ times the expected squared-norm of the honest gradients, while highly ``curved'' trajectories (e.g., close to a local minimum) tend to make $s_u$ negative.

This observation stresses that this formulation of momentum can sometimes be harmful for the purpose of Byzantine resilience.
We measured $s_u$ for every step $u\!>\!0$ in our experiments, and we always observed that this quantity is positive and increases for a short window of (dozen) steps (depending on $\lr{t}$), and then oscillates between positive and negative values.
These two phases are noticeable in the first steps of Figure \ref{fig:experiments-ratio}.
While the empirical impact (decreased or cancelled loss in accuracy) is concrete, we believe there is room for further improvements, discussed in Section \ref{sec:conclusion}.

The purpose of using momentum at the workers is to reduce the variance-norm ratio $r_t^{(w)}$, compared to $r_t^{(s)}$.
Since $\gradcomp{i}{0} = \gradsent{i}{0}$, we verify that $r_0^{(u)} = r_0^{(w)}$.
Then, $\forall t > 0$:
\begin{align}
    r_t^{(w)} \le r_t^{(s)} \Leftrightarrow{} &
    \frac{\gradsdev{t}^2 + \mu^2 \, \mntmgradsdev{t - 1}^2}{\gradnorm{t}^2 + s_t + \mu^2 \, \mntmgradnorm{t - 1}^2}
    \le \frac{\gradsdev{t}^2}{\gradnorm{t}^2} \nonumber\\
    \Leftrightarrow{} &
    \mu^2 \, \mntmgradsdev{t - 1}^2 \, \gradnorm{t}^2
    \le \left( s_t + \mu^{2} \, \mntmgradnorm{t - 1}^2 \right) \gradsdev{t}^2 \nonumber\\
    \Leftrightarrow{} &
    s_t \ge \mu^2 \, \mntmgradnorm{t - 1}^2 \left( \frac{r_{t-1}^{(w)}}{r_t^{(s)}} - 1 \right)
    \label{eq:momentum-decrease-condition}
\end{align}
The condition for decreasing $r_t^{(w)}$ can be obtained similarly:
\begin{equation*}
    r_t^{(w)} \le r_{t-1}^{(w)}
    \Leftrightarrow{}
    s_t \ge \gradnorm{t}^2 \left( \frac{r_t^{(s)}}{r_{t-1}^{(w)}} - 1 \right)
\end{equation*}

To study the impact of a lower learning rate $\lr{t}$ on $s_t$, we will assume that the real gradient $ \nabla{}\reallossletter{} $ is $l$-Lipschitz.
Namely:
\begin{align*}
    \forall \left( t, u \right) \in \setn^2, u < t,
    \norm{\expect\graddist{t} - \expect\graddist{u}}^2
    &\le l^2 \norm{\params{t} - \params{u}} \\
    &\le l^2 \norm{\sum_{v = u}^{t - 1}{\lr{v} \, \gradsent{}{v}}}
\end{align*}
Then, $\forall \left( t, u \right) \in \setn^2, u < t$, we can rewrite:
\begin{equation*}
    \norm{\expect\graddist{t} - \expect\graddist{u}}^2 =
    \underbrace{\norm{\expect\graddist{t}}^2}_{\gradnorm{t}^2} +
    \underbrace{\norm{\expect\graddist{u}}^2}_{\gradnorm{u}^2} -
    2 \, \expect\graddist{t} \cdot \expect\graddist{u}
\end{equation*}
And finally, we can lower-bound $s_t$ as:
\begin{align}
    &\sum\limits_{u = 0}^{t - 1}{\mu^{t - u} \norm{\expect\graddist{t} - \expect\graddist{u}}^2} \nonumber\\
    =\, &\sum\limits_{u = 0}^{t - 1}{\mu^{t - u} \left( \gradnorm{t}^2 + \gradnorm{u}^2 \right)}
    - \underbrace{2 \sum\limits_{u = 0}^{t - 1}{\mu^{t - u} \expect\graddist{t} \cdot \expect\graddist{u}}}_{s_t} \nonumber\\
    \le\, &\sum\limits_{u = 0}^{t - 1}{\mu^{t - u} \, l^2 \norm{\sum_{v = u}^{t - 1}{\lr{v} \, \gradsent{}{v}}}} \nonumber\\
    \Leftrightarrow{} s_t \ge\, &\sum\limits_{u = 0}^{t - 1}{\mu^{t - u} \left( \gradnorm{t}^2 + \gradnorm{u}^2 - l^2 \norm{\sum_{v = u}^{t - 1}{\lr{v} \, \gradsent{}{v}}} \right)} \label{eq:momentum-inequality}\\
    \ge\, &\frac{1 - \mu^t}{1 - \mu} \gradnorm{t}^2 + \sum\limits_{u = 0}^{t - 1}{\mu^{t - u} \left( \gradnorm{u}^2 - l^2 \norm{\sum_{v = u}^{t - 1}{\lr{v} \, \gradsent{}{v}}} \right)} \nonumber
\end{align}
When the real gradient $ \nabla{}\reallossletter{} $ is (locally) Lipschitz continuous, reducing the learning rate $\lr{t}$ can suffice to ensure $s_t$ satisfies the conditions laid above for decreasing the variance-norm ratio $r_t^{(w)}$; the purpose of momentum at the workers.

Importantly this last lower bound, namely Equation \eqref{eq:momentum-inequality}, sets how the practitioner should choose two hyperparameters, $\mu$ and $\lr{t}$, for the purpose of Byzantine-resilience.
Basically, and as long as it does not harm the training without adversary, $\mu$ should be set \emph{as high} and $\lr{t}$ \emph{as low} as possible.

As a side note, \cite{empire} is prone to increasing the lower bound on $s_t$.
Indeed, this attack submits gradients smaller or opposed to the honest gradient (Section \ref{sec:background-attack-empire}).
Such an attack can shorten the parameter trajectory, and so can improve Byzantine-resilience in the ensuing step(s).

\section{Experiments}
\label{sec:experiments}

The goal of this section is to empirically verify our theoretical results, measuring the evolution of both the top-1 cross-accuracy and variance-norm ratio over the training.
Our experiments cover every possible combinations of 5 key hyperparameters, including combinations used by \cite{little, empire}.
For reproducibility and confidence in the results, each combination of hyperparameters is repeated 5 times with \emph{seeds} 1 to 5, totalling 440 different runs.
Besides observing the benefit of lower learning rates, our results show tangible mitigation of both attacks.

\subsection{Experimental Setup}
\label{sec:experiments-setup}

We use a compact notation to define the models: L(\#outputs) for a fully-connected linear layer, R for ReLU activation, S for log-softmax, C(\#channels) for a fully-connected 2D-con\-volutional layer (kernel size 3, padding 1, stride 1), M for 2D-maxpool (kernel size 2), N for batch-normalization, and D for dropout (with fixed probability $0.25$).

We use the model and dataset from \cite{little}:\\[1mm]
\begin{tabular}{l l}
    \textbf{Model} & (784)-L(100)-R-L(10)-R-S \\
    \textbf{Dataset} & MNIST ~ ($83$ training points/gradient) \\
    \textbf{\#workers} & $n = 51$ ~ $f \in \left\lbrace 24, 12 \right\rbrace$
\end{tabular}

We also use the model and dataset from \cite{empire}:\\[1mm]
\begin{tabular}{l l}
    \textbf{Model} & (3, 32$\times$32)-C(64)-R-B-C(64)-R-B-M-D- \\
    & -C(128)-R-B-C(128)-R-B-M-D- \\
    & -L(128)-R-D-L(10)-S \\
    \textbf{Dataset} & CIFAR-10 ~ ($50$ training points/gradient) \\
    \textbf{\#workers} & $n = 25$ ~ $f \in \left\lbrace 11, 5 \right\rbrace$
\end{tabular}

For model training, we use the \emph{negative log likelihood} loss and respectively $10^{-4}$ and $10^{-2}$ $\ell{}_2$-regularization for the MNIST and CIFAR-10 models.
We also clip gradients, ensuring their norms remain respectively below $2$ and $5$ for the MNIST and CIFAR-10 models.
For model evaluation, we use the \emph{top-1 cross-accuracy} on the whole testing set.

Both datasets are pre-processed before training.
For MNIST we apply the same pre-processing as in \cite{little}: an input image normalization with mean $0.1307$ and standard deviation $0.3081$.
For CIFAR-10, besides including horizontal flips of the input pictures, we also apply a per-channel normalization with means $0.4914, 0.4822, 0.4465$ and standard deviations $0.2023, 0.1994, 0.2010$ \cite{cifar10-transforms}.

We set $f$ the number of Byzantine workers either to the maximum for which \krum{} can be used (roughly an half: $f = \floor{\frac{n - 3}{2}}$), or the maximum for \bulyan{} (roughly a quarter, $f = \floor{\frac{n - 3}{4}}$).
The attack factors $\varepsilon_t$ (Section \ref{sec:background-attack}) are set to constants proposed in the literature, namely $\varepsilon_t = 1.5$ for \cite{little} and $\varepsilon_t = 1.1$ for \cite{empire}.

Guided by our theoretical study on the impact of the learning rate on the variance-norm ratio,
for every pair model-attack in our experiments we select two different learning rates.
The first and largest is selected so as to maximize the \emph{performance} (highest final cross-accuracy and accuracy gain per step) of the model trained without Byzantine workers.
The second and smallest is chosen so as to minimize the \emph{performance loss} under attack, without substantially impacting the final accuracy when trained without Byzantine workers.

The MNIST and CIFAR-10 model are trained respectively with $\mu = 0.9$ and $\mu = 0.99$.
These values were obtained by trial and error, to maximize overall accuracy gain per step.

Our theoretical analysis highlights two metrics: the \emph{top-1 cross-accuracy}, measuring the performance of the model, and the \emph{variance-norm ratio}, i.e.\ either $r_t^{(s)}$ or $r_t^{(w)}$ in accordance with where momentum was carried out.
Each experiment is run 5 times.
We present the average and standard deviation of the two metrics over these 5 runs.

\subsection{Reproducibility}

Particular care has been taken to make our results reproducible.
Each of the 5 runs per experiment are respectively seeded with seed 1 to 5.
For instance, this implies that two experiments with same seed and same model also starts with the same parameters $\params{0}$.
To further reduce the sources of non-determinism, the CuDNN backend is configured in deterministic mode (our experiments ran on a \emph{GeForce GTX 1080 Ti}) with benchmark mode turned off.
We also used \emph{log-softmax} + \emph{nll loss}, which is equal to \emph{softmax} + \emph{cross-entropy loss}, but with improved numerical stability on PyTorch.

We provide our code along with a script reproducing \emph{all} of our results, both the experiments and the graphs, in one command.
Details, including software and hardware dependencies, are available in the supplementary material.

\subsection{Experimental Results}
\label{sec:experiments-results}

\begin{figure}[p!]
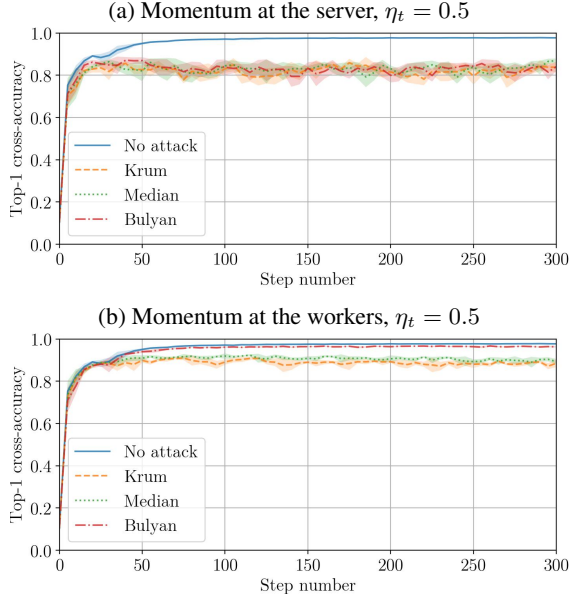

    \centering
    \begin{subfigure}[t]{0.9\linewidth}
        \centering
        \multicaption{Momentum at the server, $\lr{t} = 0.5$}
		\includeplot{\detokenize{mnist-little-f_12-lr_0.5-at_update}}%
		\label{fig:experiments-mnist-little-a}
	\end{subfigure}
	\begin{subfigure}[t]{0.9\linewidth}
		\centering
		\multicaption{Momentum at the workers, $\lr{t} = 0.5$}
		\includeplot{\detokenize{mnist-little-f_12-lr_0.5-at_worker}}%
		\label{fig:experiments-mnist-little-b}
	\end{subfigure}
    \caption{MNIST and the associated model (Section \ref{sec:experiments-setup}), $n = 51$ and $f = 12$.
    The attack, \cite{little}, has a tangible impact ($-15\%$ on the maximum accuracy) in this setup.
    Using momentum at the workers diminishes the effect of the attack, even substantially when \bulyan{} is used (only $1\%$ loss in accuracy).}
    \label{fig:experiments-first}
    \label{fig:experiments-mnist-little}
\end{figure}

\begin{figure}[p!]
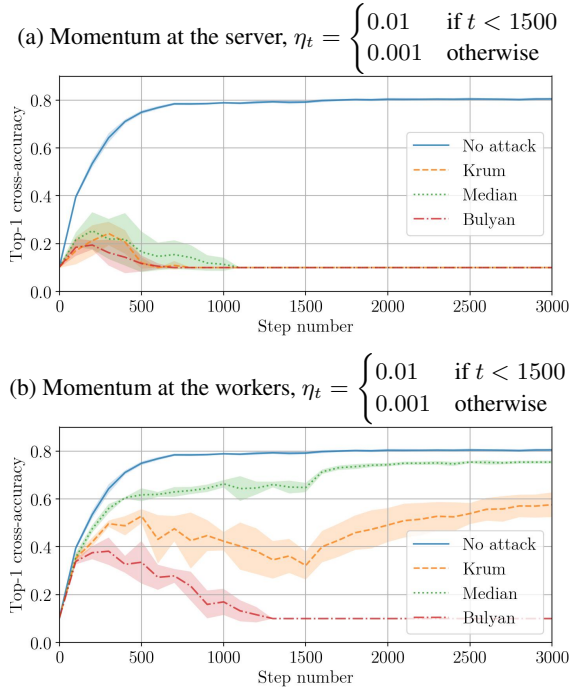

    \centering
    \begin{subfigure}[t]{0.9\linewidth}
        \centering
        \multicaption{Momentum at the server, \lrtitle{}}
		\includeplot{\detokenize{cifar10-little-f_5-lr_0.01-at_update}}%
		\label{fig:experiments-cifar10-little-a}
	\end{subfigure}
	\begin{subfigure}[t]{0.9\linewidth}
		\centering
		\multicaption{Momentum at the workers, \lrtitle{}}
		\includeplot{\detokenize{cifar10-little-f_5-lr_0.01-at_worker}}%
		\label{fig:experiments-cifar10-little-b}
	\end{subfigure}
    \caption{CIFAR-10 and the associated model (Section \ref{sec:experiments-setup}), with $n = 25$ and $f = 5$.
    As in Figure \ref{fig:experiments-mnist-little}, \cite{little} has a strong impact on the training.
    The positive effect of momentum at the workers (Figure \ref{fig:experiments-cifar10-little-b}) is conspicuous and, as predicted by the theory (Section \ref{sec:momentum-effects}), amplified with a lower learning rate.}
    \label{fig:experiments-cifar10-little}
\end{figure}

\begin{figure}[p!]
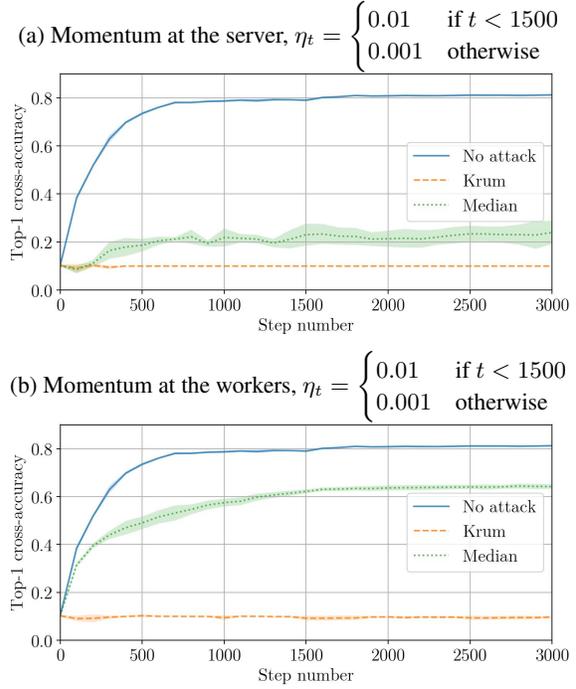

    \centering
    \begin{subfigure}[t]{0.9\linewidth}
        \centering
        \multicaption{Momentum at the server, \lrtitle{}}
		\includeplot{\detokenize{cifar10-empire-f_11-lr_0.01-at_update}}%
		\label{fig:experiments-cifar10-empire-a}
	\end{subfigure}
	\begin{subfigure}[t]{0.9\linewidth}
		\centering
		\multicaption{Momentum at the workers, \lrtitle{}}
		\includeplot{\detokenize{cifar10-empire-f_11-lr_0.01-at_worker}}%
		\label{fig:experiments-cifar10-empire-b}
	\end{subfigure}
    \caption{CIFAR-10 and the associated model (Section \ref{sec:experiments-setup}), with $n = 25$ and $f = 11$ (Figure \ref{fig:experiments-cifar10-little} uses $f = 5$).
    The attack, \cite{empire}, is extremely efficient when $f \approx \frac{n}{2}$, but had almost no effect with $f \approx \frac{n}{4}$; the supplementary material has a more complete range of experiments.
    Momentum at the workers noticeably improves the cross-accuracy when \medianoid{} is used in this setting.}
    \label{fig:experiments-cifar10-empire}
\end{figure}

\begin{figure}[p!]
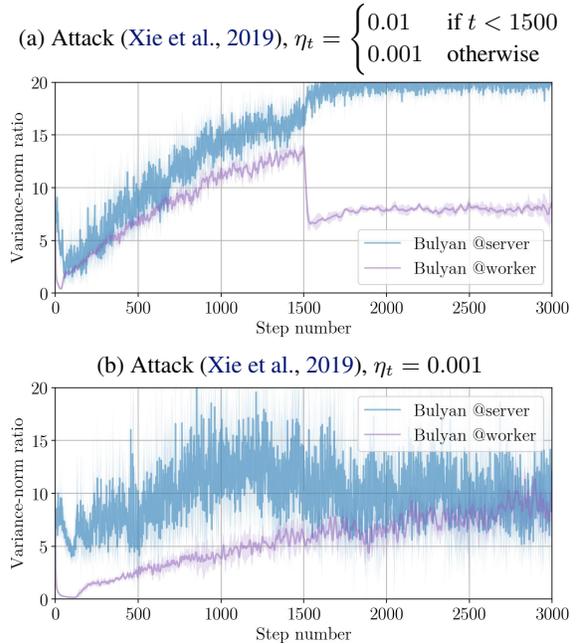

    \centering
    \begin{subfigure}[t]{0.9\linewidth}
        \centering
        \multicaption{Attack \cite{empire}, \lrtitle{}}
		\includeplot{\detokenize{cifar10-empire-bulyan-f_5-lr_0.01-ratio}}%
		\label{fig:experiments-ratio-a}
	\end{subfigure}
	\begin{subfigure}[t]{0.9\linewidth}
        \centering
        \multicaption{Attack \cite{empire}, $\lr{t} = 0.001$}
		\includeplot{\detokenize{cifar10-empire-bulyan-f_5-lr_0.001-ratio}}%
		\label{fig:experiments-ratio-b}
	\end{subfigure}
    \caption{CIFAR-10 and the associated model (Section \ref{sec:experiments-setup}), with $n = 25$ and $f = 5$, showing a decreased variance-norm ratio when momentum is computed at the workers.
    The benefit of a lower $\lr{t}$ is clearly visible in (a), when its value is reduced at step 1500.}
    \label{fig:experiments-last}
    \label{fig:experiments-ratio}
\end{figure}

For each of the pair model-dataset, we consider 5 variable hyperparameters:
which attack to test (\cite{little} or \cite{empire}), which defense to run (\krum{}, \medianoid{} or \bulyan{}), how many Byzantine workers $f$ to use (\emph{an half} or \emph{a quarter}), where momentum is computed (\emph{at the server} or \emph{at the workers}) and which learning rate $\lr{t}$ to apply.

We report on every possible combination of these hyperparameters, along with baselines that use \emph{averaging} without attack.
With 5 repetitions per setup, the experiments consist in 440 runs, aggregated and studied in the supplementary material.
In this section we report on a representative subset.

We made a concerning observation: one of the theoretical requirement for Byzantine-resilience, equations \eqref{eq:background-ratio-bound} or \eqref{eq:background-ratio-bound-median}, is actually rarely satisfied in practice.
In less than 2\% of the runs under attack was this theoretical condition satisfied for at least 1 step, and none for more than 4 steps (out of 3000).
In most of the experiments, the observed variance-norm ratio was often between 1 and 2 \emph{orders of magnitude} too high (e.g., Figure \ref{fig:experiments-ratio}).
Since our hyperparameters (model, dataset, mini-batch size, $n$, $f$) are very close, if not equal, to those used in the experiments of \cite{little, empire}, such a substantial margin (1 to 2 orders of magnitude) lets us think the theoretical requirements for Byzantine resilience were actually not satisfied either in \cite{little, empire}.
\cite{little} reached the same conclusion, using a different experiment.

\vspace*{-0.2cm}
\paragraph{Result Analysis.}
With the largest, optimal learning rate, we obtained with \krum{} and for both models very similar maximum top-1 cross-accuracies to the ones obtained in \cite{little}\footnote{Although \cite{little} uses \krum{} with $m = 1$ (see Section \ref{sec:background-gar-krum}), and not the exact same CIFAR-10 model.}, namely $\smallsim{}80\%$ for MNIST's model (Figure \ref{fig:experiments-mnist-little-a}) and $\smallsim{}20\%$ for CIFAR-10's model (Figure \ref{fig:experiments-cifar10-little-a}).

A similar observation can be made for \cite{empire}: on the CIFAR-10's model, the maximum accuracies of \krum{} and \medianoid{} are respectively $\smallsim{}10\%$ and $\smallsim{}20\%$ (Figure \ref{fig:experiments-cifar10-empire-a}).

The benefit of computing momentum at the workers is visible in all the figures.
Regarding the impact on the top-1 cross-accuracy, we systematically\footnote{Except for \krum{} against \cite{empire} when $f = \floor{\frac{n - 3}{2}}$.} observe an increase compared to when momentum is computed at the server (figures \ref{fig:experiments-mnist-little}--\ref{fig:experiments-cifar10-empire}).
The empirical increase ranges from $+5\%$, on the MNIST model attacked by \cite{little} (supplementary material, Figure 3), to $+50\%$ on the CIFAR-10 model, defended by \medianoid{} (Figure \ref{fig:experiments-cifar10-little}).
Regarding the effect on the variance-norm ratio, we comparatively observe a decrease of this ratio before approaching convergence.
As predicted by our theoretical analysis, this decrease can be amplified by reducing the learning rate (Figure \ref{fig:experiments-ratio-a}).

\section{Concluding Remarks}
\label{sec:conclusion}

\paragraph{Momentum-based Variance Reduction.}
Our algorithm is different from \cite{momentum-variance}, as instead of reducing the variance of the gradients, we actually \emph{increase} it (Equation \eqref{eq:momentum-variance}).
What we seek to reduce is the \emph{variance-norm ratio}, which is the key quantity for any Byzantine-resilient GAR approximating a high-dimensional median, e.g.\ \krum{}, \medianoid{}, \bulyan{} as well as in \cite{yang2019byrdie, yang2019bridge, geomed, munoz2019byzantine}\footnote{This list is not exhaustive.}.

Some of the ideas introduced in \cite{momentum-variance} could nevertheless help further improve Byzantine resilience.
For instance, introducing an adaptive learning rate which decreases depending on the curvature of the parameter trajectory is an appealing approach to further reduce the variance-norm ratio (Equation \eqref{eq:momentum-inequality}).

\paragraph{Further Work.}
The theoretical condition for ratio reduction, in Section \ref{sec:momentum-effects}, shows that momentum at the workers is a double-edged sword.
The intuition can be gained with the classic analogy from physics:
without Byzantine workers, momentum makes the parameters $\params{t}$ somehow like a particle travelling down the loss function with \emph{inertia}.
When inside a ``straight valley'', past estimation errors are on average compensated in the next steps, dampening oscillations and accumulating the average descent direction.
The variance-norm ratio of the momentum gradient is then reduced, mostly because its norm \emph{increases}, which is quantified by $s_t$.
The problem is that $s_t$ can become negative.
Intuitively with the particle analogy, this happens when the loss is locally ``curved'', for instance when approaching a local minimum.
The particle may start continuing ``uphill'' instead of following the ``valley'', and so, losing \emph{momentum}.
The norm of the momentum gradient then \emph{decreases}, increasing the variance-norm ratio.

While the ability to cross narrow, local minima is recognized as an accelerator \cite{momentum-why}, for the purpose of Byzantine-resilience we want to ensure momentum at the workers does not increase the variance-norm ratio compared to the classical, momentum at the server.
The theoretical condition for this purpose is given in Equation \eqref{eq:momentum-decrease-condition}.
One simple amendment would then be to use momentum at the workers when Equation \eqref{eq:momentum-decrease-condition} is satisfied, and fallback to computing it at the server otherwise.
Also, a more complex, possible future approach could be to dynamically adapt the momentum factor $\mu$, decreasing it as the curvature increases.

\paragraph{Asynchronous SGD.} We focused in this work on the synchronous setting, which received most of the attention in the Byzantine-resilient literature.
Yet, we believe our work can be applied to asynchronous settings, as momentum  is agnostic to the question of synchrony.
Specifically, combining our idea with a filtering scheme such as \kardam~\cite{kardam} is in principle possible, as the filter commutes with the basic operations of momentum.
However, further analysis of the interplay between the dynamics of stale gradients and the dynamics of momentum remain necessary.

\paragraph{Byzantine Servers.} While most of the research on Byzantine-resilience gradient descent has focused on the workers' side, 
assuming a reliable server, recent efforts have started tackling Byzantine servers~\cite{guanyu}. Our reduction of the variance-norm ratio  
strengthens the gradient aggregation phase, which is necessary whether we deal with Byzantine workers or Byzantine servers. An interesting open question is how the momentum dynamics affects the models drift between different parameter servers. Any quantitative answer to this question will enable the use of our method in fully decentralised Byzantine resilient gradient descent.



\bibliography{bib}
\bibliographystyle{style/icml2019}

\appendix
\clearpage
\section{Reproducing the results}

The codebase is available at \url{https://github.com/LPD-EPFL/ByzantineMomentum}.

\subsection{Dependencies}

\paragraph{Software dependencies.}
Python 3.7.3 has been used to run our scripts.
Besides the standard libraries associated with Python 3.7.3, our scripts also depend on:

\hfill
\begin{tabular}{l l}
    \hline
    \textbf{Library} & \textbf{Version} \\
    \hline
    numpy & 1.17.2 \\
    torch & 1.2.0 \\
    torchvision & 0.4.0 \\
    pandas & 0.25.1 \\
    matplotlib & 3.0.2 \\
    tqdm & 4.40.2 \\
    PIL & 6.1.0 \\
    \hline
\end{tabular}
\hfill
\begin{tabular}{l l}
    \hline
    \textbf{Library} & \textbf{Version} \\
    \hline
    six & 1.12.0 \\
    pytz & 2019.3 \\
    dateutil & 2.7.3 \\
    pyparsing & 2.2.0 \\
    cycler & 0.10.0 \\
    kiwisolver & 1.0.1 \\
    cffi & 1.13.2 \\
    \hline
\end{tabular}
\hfill\null

We list below the OS on which our scripts have been tested:
\begin{itemize}
    \item{Debian 10 (GNU/Linux 4.19.0-6 x86\_64)}
    \item{Ubuntu 18.04.3 LTS (GNU/Linux 4.15.0-58 x86\_64)}
\end{itemize}

\paragraph{Hardware dependencies.}
Although our experiments are time-agnostic, we list below the hardware components used:
\begin{itemize}
    \item{1 ~ Intel(R) Core(TM) i7-8700K CPU @ 3.70GHz}
    \item{2 ~ Nvidia GeForce GTX 1080 Ti}
    \item{64 GB of RAM}
\end{itemize}

\subsection{Command}

Our results, i.e.\ the experiments and graphs, are reproducible in one command.
In the root directory, please run:

\hfill\colorbox{gray!20}{\parbox[c][5mm]{0.8\linewidth}{\texttt{\$ python3 reproduce.py}}}\hfill\null

On our hardware, reproducing the results takes $\smallsim{}$24 hours.

\newpage
\section{Experimental results}

For every pair model-dataset, the following parameters vary:
\begin{itemize}
    \item{Which attack: \cite{little} or \cite{empire}}
    \item{Which defense: \krum{}, \medianoid{} or \bulyan{}}
    \item{How many Byzantine workers (\emph{an half} or \emph{a quarter})}
    \item{Where momentum is computed (\emph{server} or \emph{workers})}
    \item{Which learning rate is used (\emph{larger} or \emph{smaller})}
\end{itemize}
Every possible combination is tested\footnote{Along with baselines using \emph{averaging} without attack.}, leading to a total of 88 different experiment setups.
Each setup is tested 5 times, each run with a fixed seed from 1 to 5, enabling verbatim reproduction of our results\footnote{Despite our best efforts, there may still exist minor sources of non-determinism, like race-conditions in the evaluation of certain functions (e.g., parallel additions) in a GPU.
Nevertheless we believe these should not affect the results in any significant way.}.
We then report the average and standard deviation for two metrics:
\emph{top-1 cross-accuracy} and \emph{variance-norm ratio} over the training steps.

\begin{figure*}
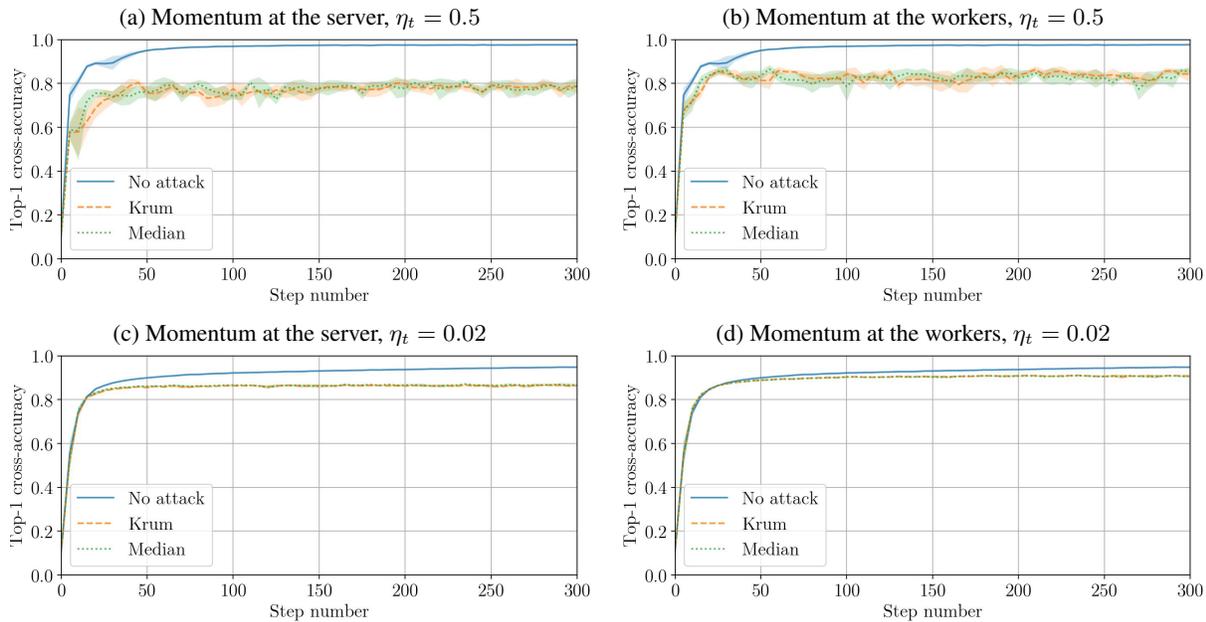

    \centering
    \begin{subfigure}[t]{0.45\linewidth}
        \centering
        \multicaption{Momentum at the server, $\lr{t} = 0.5$}
		\includeplot{\detokenize{mnist-little-f_24-lr_0.5-at_update}}
	\end{subfigure}
	\quad
	\begin{subfigure}[t]{0.45\linewidth}
		\centering
		\multicaption{Momentum at the workers, $\lr{t} = 0.5$}
		\includeplot{\detokenize{mnist-little-f_24-lr_0.5-at_worker}}
	\end{subfigure}
	\begin{subfigure}[t]{0.45\linewidth}
        \centering
        \multicaption{Momentum at the server, $\lr{t} = 0.02$}
		\includeplot{\detokenize{mnist-little-f_24-lr_0.02-at_update}}
	\end{subfigure}
	\quad
	\begin{subfigure}[t]{0.45\linewidth}
		\centering
		\multicaption{Momentum at the workers, $\lr{t} = 0.02$}
		\includeplot{\detokenize{mnist-little-f_24-lr_0.02-at_worker}}
	\end{subfigure}
    \caption{MNIST using $n = 51$ workers, including $f = 24$ Byzantine workers implementing \cite{little}.
    This is the maximum number of Byzantine workers $\krum{}$ can support.
    No matter where the momentum is computed, reducing the learning rate decreases the effect of the attack against all the GARs.
    This is not observed, in the same setting, when \cite{empire} is used instead (Figure \ref{fig:xacc-mnist-empire-half}).
    No matter the learning rate, using momentum at the workers always leads in these settings to an increase of the final accuracy (+5\%).}
    \label{fig:xacc-mnist-little-half}
    \label{fig:xacc-first}
\end{figure*}

\begin{figure*}
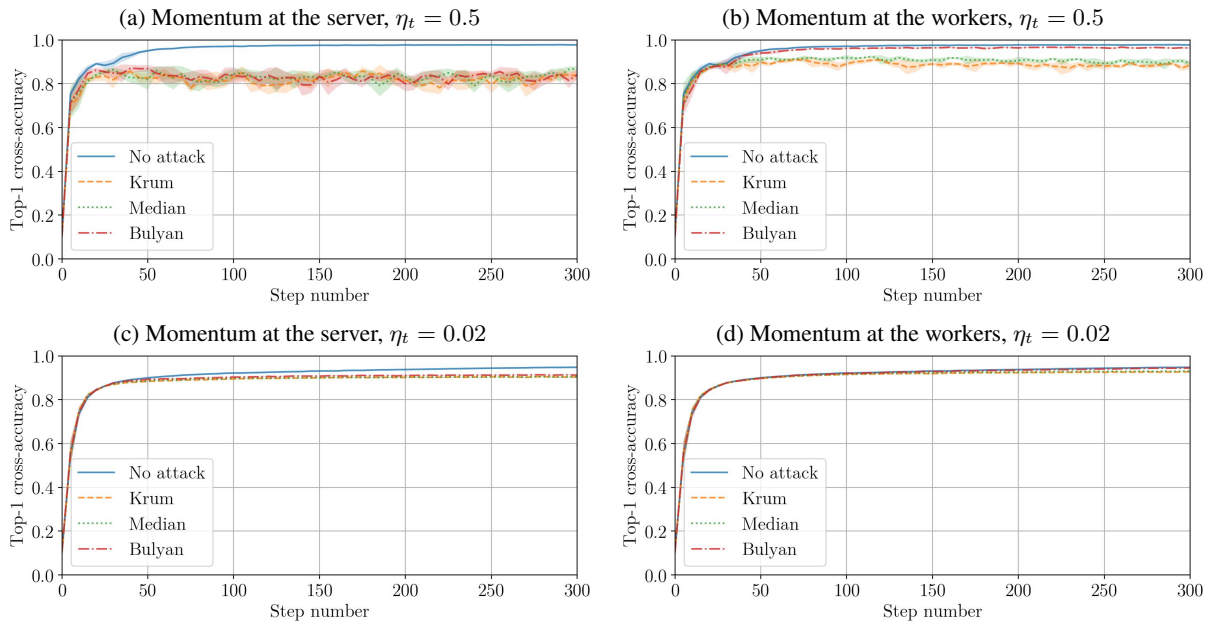

    \centering
    \begin{subfigure}[t]{0.45\linewidth}
        \centering
        \multicaption{Momentum at the server, $\lr{t} = 0.5$}
		\includeplot{\detokenize{mnist-little-f_12-lr_0.5-at_update}}
	\end{subfigure}
	\quad
	\begin{subfigure}[t]{0.45\linewidth}
		\centering
		\multicaption{Momentum at the workers, $\lr{t} = 0.5$}
		\includeplot{\detokenize{mnist-little-f_12-lr_0.5-at_worker}}
	\end{subfigure}
	\begin{subfigure}[t]{0.45\linewidth}
        \centering
        \multicaption{Momentum at the server, $\lr{t} = 0.02$}
		\includeplot{\detokenize{mnist-little-f_12-lr_0.02-at_update}}
	\end{subfigure}
	\quad
	\begin{subfigure}[t]{0.45\linewidth}
		\centering
		\multicaption{Momentum at the workers, $\lr{t} = 0.02$}
		\includeplot{\detokenize{mnist-little-f_12-lr_0.02-at_worker}}
	\end{subfigure}
    \caption{MNIST using $n = 51$ workers, including $f = 12$ Byzantine workers implementing \cite{little}.
    This is the maximum number of Byzantine workers $\bulyan{}$ can support.
    With momentum at the server and the largest learning rate, the impact of the attack remains unchanged compared to Figure \ref{fig:xacc-mnist-little-half} and despite the reduced number of Byzantine workers.
    \bulyan{}, which combines \krum{} and \medianoid{}, achieves the same performance as both \krum{} and \medianoid{}.
    When momentum is computed at the workers, \bulyan{} achieves in these settings better resilience than its parts \krum{} and \medianoid{}, and the attack has no tangible effect anymore.}
    \label{fig:xacc-mnist-little-quarter}
\end{figure*}

\begin{figure*}
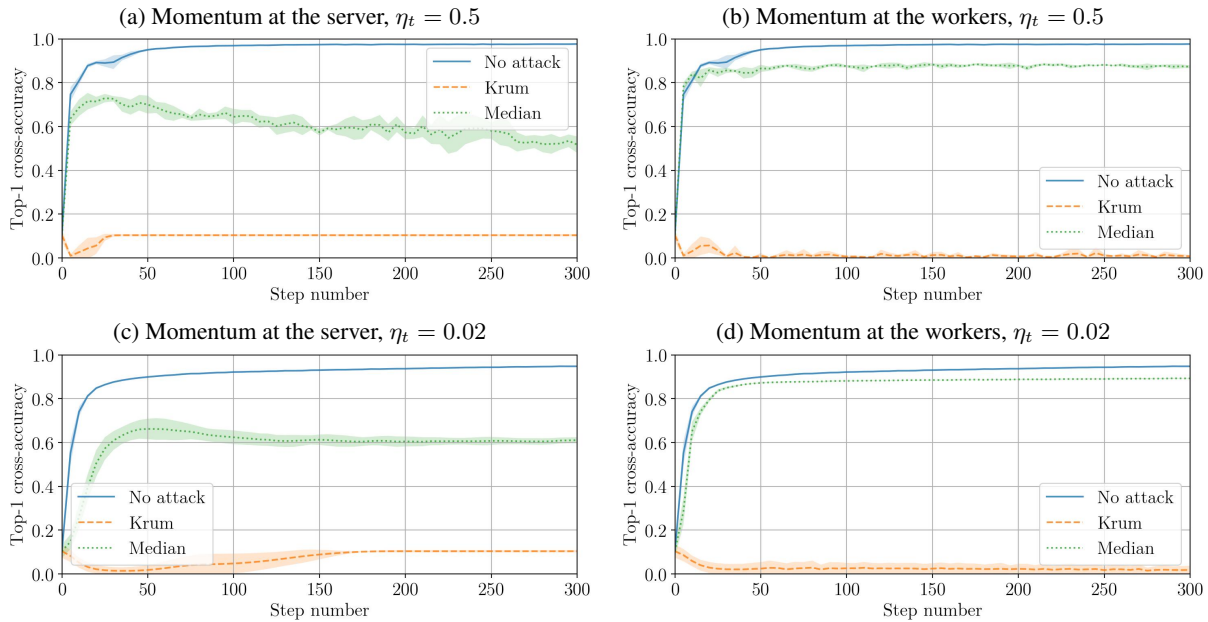

    \centering
    \begin{subfigure}[t]{0.45\linewidth}
        \centering
        \multicaption{Momentum at the server, $\lr{t} = 0.5$}
		\includeplot{\detokenize{mnist-empire-f_24-lr_0.5-at_update}}
	\end{subfigure}
	\quad
	\begin{subfigure}[t]{0.45\linewidth}
		\centering
		\multicaption{Momentum at the workers, $\lr{t} = 0.5$}
		\includeplot{\detokenize{mnist-empire-f_24-lr_0.5-at_worker}}
	\end{subfigure}
	\begin{subfigure}[t]{0.45\linewidth}
        \centering
        \multicaption{Momentum at the server, $\lr{t} = 0.02$}
		\includeplot{\detokenize{mnist-empire-f_24-lr_0.02-at_update}}
	\end{subfigure}
	\quad
	\begin{subfigure}[t]{0.45\linewidth}
		\centering
		\multicaption{Momentum at the workers, $\lr{t} = 0.02$}
		\includeplot{\detokenize{mnist-empire-f_24-lr_0.02-at_worker}}
	\end{subfigure}
    \caption{MNIST using $n = 51$ workers, including $f = 24$ Byzantine workers implementing \cite{empire}.
    This is the maximum number of Byzantine workers $\krum{}$ can support.
    Contrary to Figure \ref{fig:xacc-mnist-little-half}, this attack has very different impacts on the training depending on the Byzantine-resilient GAR used.
    With \krum{} and momentum at the server, the model parameters quickly (after 30 steps) reach a point where the output class becomes independent from the input; the model is driven useless.
    Momentum at the workers with \krum{} only avoids obtaining such a model.
    \medianoid{} shows substantially more resilience than \krum{} in this setup, and when momentum is computed at the workers, the maximum cross-accuracy is consistently increased, between 15\% to 25\% additional points.}
    \label{fig:xacc-mnist-empire-half}
\end{figure*}

\begin{figure*}
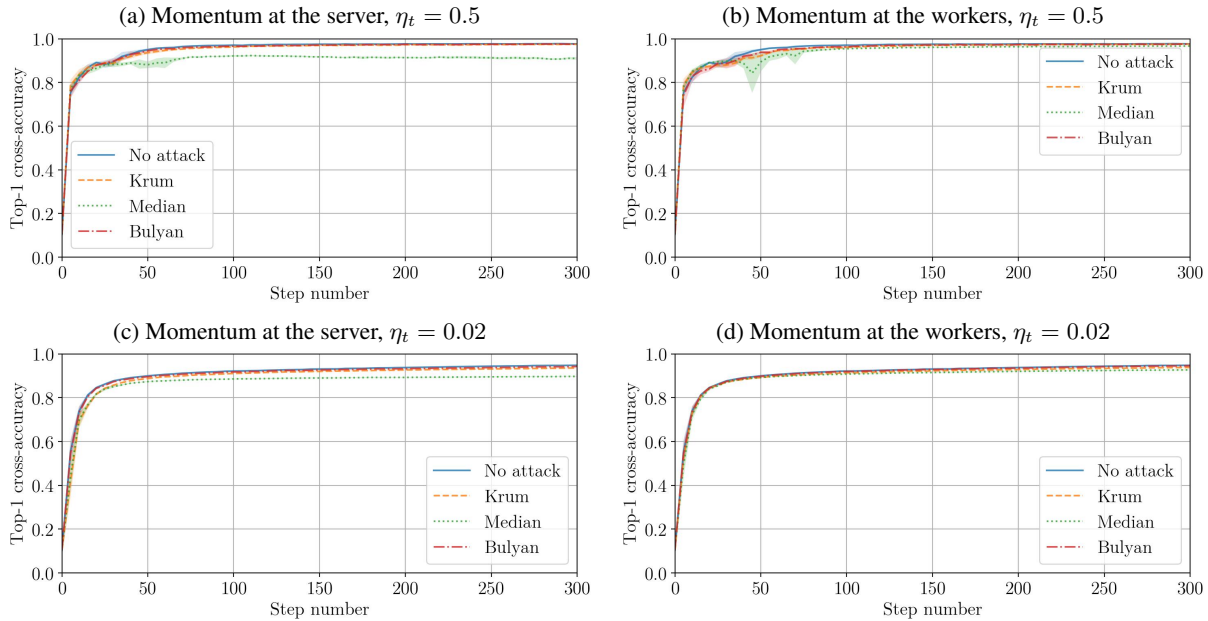

    \centering
    \begin{subfigure}[t]{0.45\linewidth}
        \centering
        \multicaption{Momentum at the server, $\lr{t} = 0.5$}
		\includeplot{\detokenize{mnist-empire-f_12-lr_0.5-at_update}}
	\end{subfigure}
	\quad
	\begin{subfigure}[t]{0.45\linewidth}
		\centering
		\multicaption{Momentum at the workers, $\lr{t} = 0.5$}
		\includeplot{\detokenize{mnist-empire-f_12-lr_0.5-at_worker}}
	\end{subfigure}
	\begin{subfigure}[t]{0.45\linewidth}
        \centering
        \multicaption{Momentum at the server, $\lr{t} = 0.02$}
		\includeplot{\detokenize{mnist-empire-f_12-lr_0.02-at_update}}
	\end{subfigure}
	\quad
	\begin{subfigure}[t]{0.45\linewidth}
		\centering
		\multicaption{Momentum at the workers, $\lr{t} = 0.02$}
		\includeplot{\detokenize{mnist-empire-f_12-lr_0.02-at_worker}}
	\end{subfigure}
    \caption{MNIST using $n = 51$ workers, including $f = 12$ Byzantine workers implementing \cite{empire}.
    This is the maximum number of Byzantine workers $\bulyan{}$ can support.
    With a quarter of Byzantine workers, the attack does not have any tangible impact on \krum{} (and thus none on \bulyan{} either) for this model and dataset anymore.
    Using momentum at the workers further reduces the impact on \medianoid{}, to the point of filtering out the adversarial effect of this attack.}
    \label{fig:xacc-mnist-empire-quarter}
\end{figure*}

\begin{figure*}
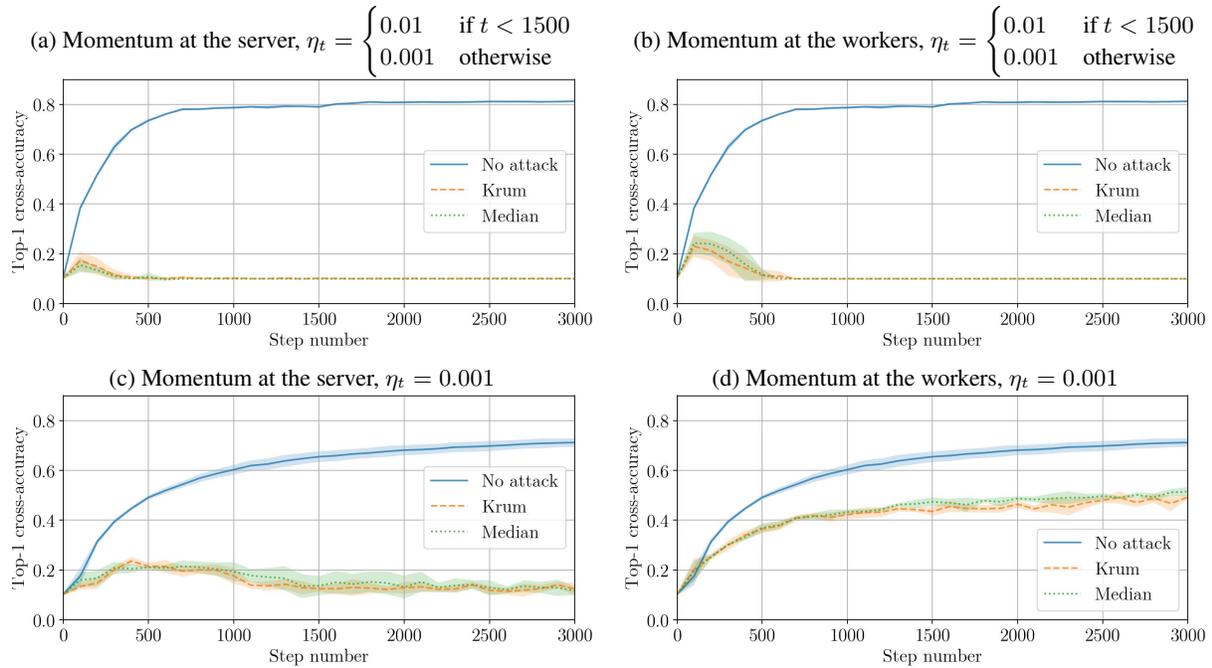

    \centering
    \begin{subfigure}[t]{0.45\linewidth}
        \centering
        \multicaption{Momentum at the server, \lrtitle{}}
		\includeplot{\detokenize{cifar10-little-f_11-lr_0.01-at_update}}
	\end{subfigure}
	\quad
	\begin{subfigure}[t]{0.45\linewidth}
		\centering
		\multicaption{Momentum at the workers, \lrtitle{}}
		\includeplot{\detokenize{cifar10-little-f_11-lr_0.01-at_worker}}
	\end{subfigure}
	\begin{subfigure}[t]{0.45\linewidth}
        \centering
        \multicaption{Momentum at the server, $\lr{t} = 0.001$}
		\includeplot{\detokenize{cifar10-little-f_11-lr_0.001-at_update}}
	\end{subfigure}
	\quad
	\begin{subfigure}[t]{0.45\linewidth}
		\centering
		\multicaption{Momentum at the workers, $\lr{t} = 0.001$}
		\includeplot{\detokenize{cifar10-little-f_11-lr_0.001-at_worker}}
	\end{subfigure}
    \caption{CIFAR-10 using $n = 25$ workers, including $f = 11$ Byzantine workers implementing \cite{little}.
    This is the maximum number of Byzantine workers $\krum{}$ can support.
    Compared to Figure \ref{fig:xacc-mnist-little-half}, the impact of the attack is substantial.
    Even with a reduced learning rate, the model maximum cross-accuracy barely reaches 20\%.
    Using momentum at the workers, while having virtually no computational cost, positively impacts the performance of the model.}
    \label{fig:xacc-cifar-little-half}
\end{figure*}

\begin{figure*}
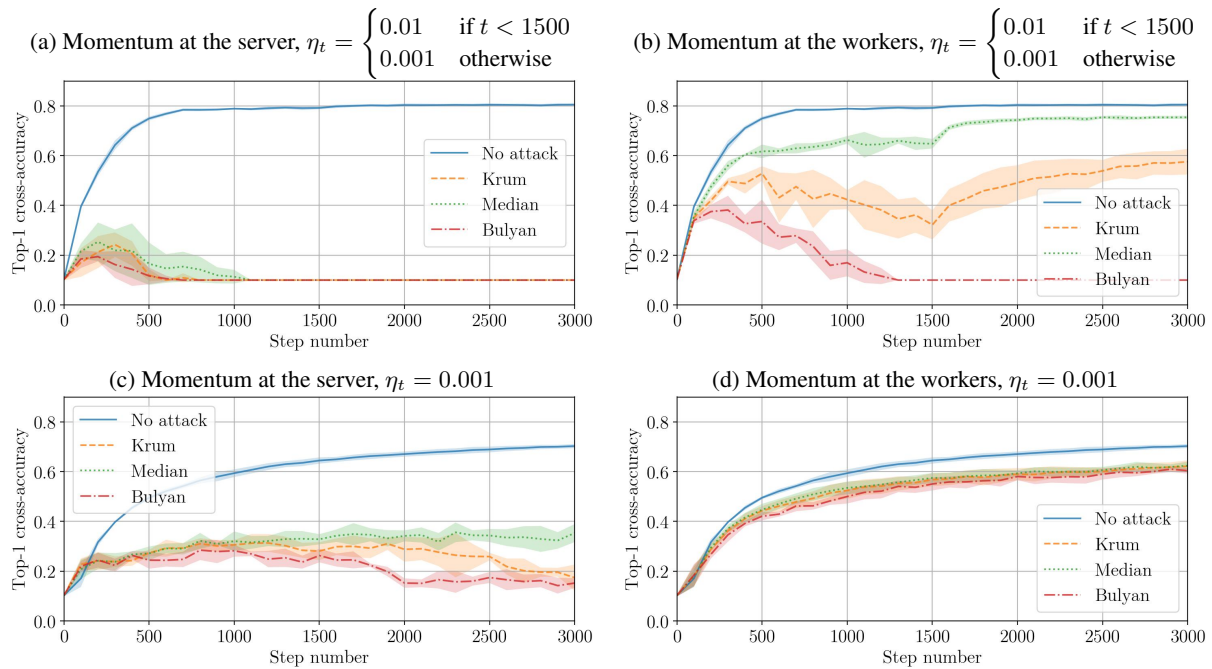

    \centering
    \begin{subfigure}[t]{0.45\linewidth}
        \centering
        \multicaption{Momentum at the server, \lrtitle{}}
		\includeplot{\detokenize{cifar10-little-f_5-lr_0.01-at_update}}
	\end{subfigure}
	\quad
	\begin{subfigure}[t]{0.45\linewidth}
		\centering
		\multicaption{Momentum at the workers, \lrtitle{}}
		\includeplot{\detokenize{cifar10-little-f_5-lr_0.01-at_worker}}
	\end{subfigure}
	\begin{subfigure}[t]{0.45\linewidth}
        \centering
        \multicaption{Momentum at the server, $\lr{t} = 0.001$}
		\includeplot{\detokenize{cifar10-little-f_5-lr_0.001-at_update}}
	\end{subfigure}
	\quad
	\begin{subfigure}[t]{0.45\linewidth}
		\centering
		\multicaption{Momentum at the workers, $\lr{t} = 0.001$}
		\includeplot{\detokenize{cifar10-little-f_5-lr_0.001-at_worker}}
	\end{subfigure}
    \caption{CIFAR-10 using $n = 25$ workers, including $f = 5$ Byzantine workers implementing \cite{little}.
    This is the maximum number of Byzantine workers $\bulyan{}$ can support.
    The effect of going to only a quarter of Byzantine workers, compared to Figure \ref{fig:xacc-cifar-little-half}, did not made the attack less effective.
    Conversely, momentum at the workers leads to a conspicuous improvement of the model performance.}
    \label{fig:xacc-cifar-little-quarter}
\end{figure*}

\begin{figure*}
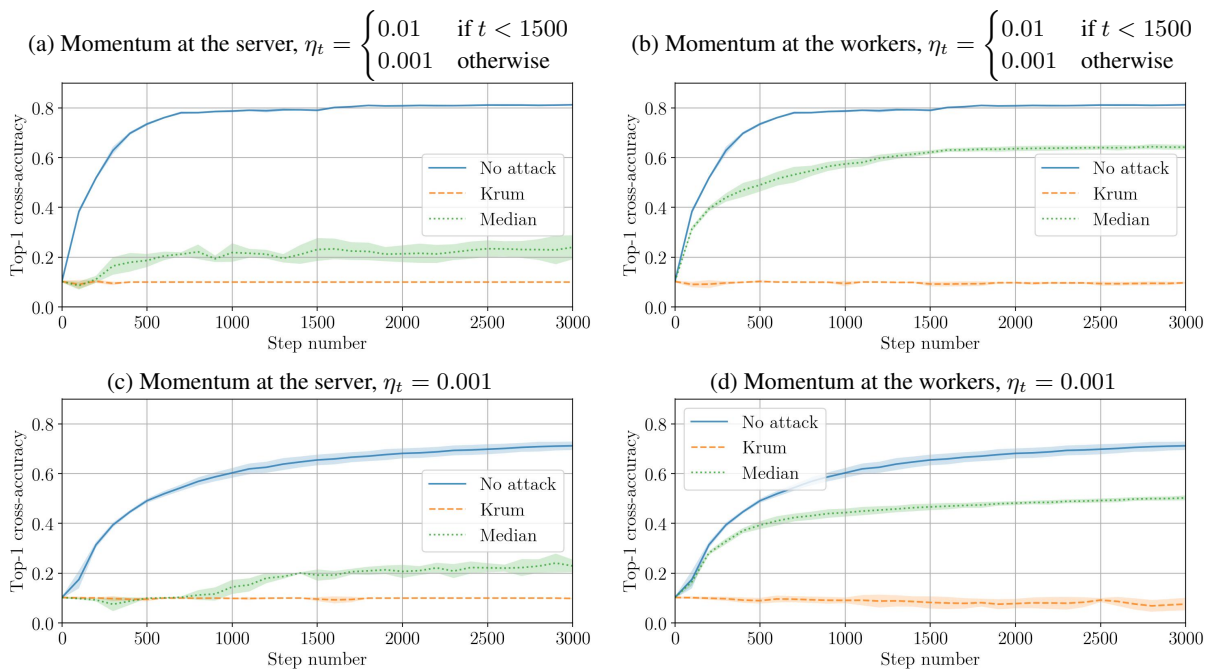

    \centering
    \begin{subfigure}[t]{0.45\linewidth}
        \centering
        \multicaption{Momentum at the server, \lrtitle{}}
		\includeplot{\detokenize{cifar10-empire-f_11-lr_0.01-at_update}}
	\end{subfigure}
	\quad
	\begin{subfigure}[t]{0.45\linewidth}
		\centering
		\multicaption{Momentum at the workers, \lrtitle{}}
		\includeplot{\detokenize{cifar10-empire-f_11-lr_0.01-at_worker}}
	\end{subfigure}
	\begin{subfigure}[t]{0.45\linewidth}
        \centering
        \multicaption{Momentum at the server, $\lr{t} = 0.001$}
		\includeplot{\detokenize{cifar10-empire-f_11-lr_0.001-at_update}}
	\end{subfigure}
	\quad
	\begin{subfigure}[t]{0.45\linewidth}
		\centering
		\multicaption{Momentum at the workers, $\lr{t} = 0.001$}
		\includeplot{\detokenize{cifar10-empire-f_11-lr_0.001-at_worker}}
	\end{subfigure}
    \caption{CIFAR-10 using $n = 25$ workers, including $f = 11$ Byzantine workers implementing \cite{empire}.
    This is the maximum number of Byzantine workers $\krum{}$ can support.
    As in Figure \ref{fig:xacc-mnist-empire-half}, the attack is extremely effective on \krum{} and slightly less on \medianoid{}.
    Momentum at the workers has a substantial positive effect when \medianoid{} is used.}
    \label{fig:xacc-cifar-empire-half}
\end{figure*}

\begin{figure*}
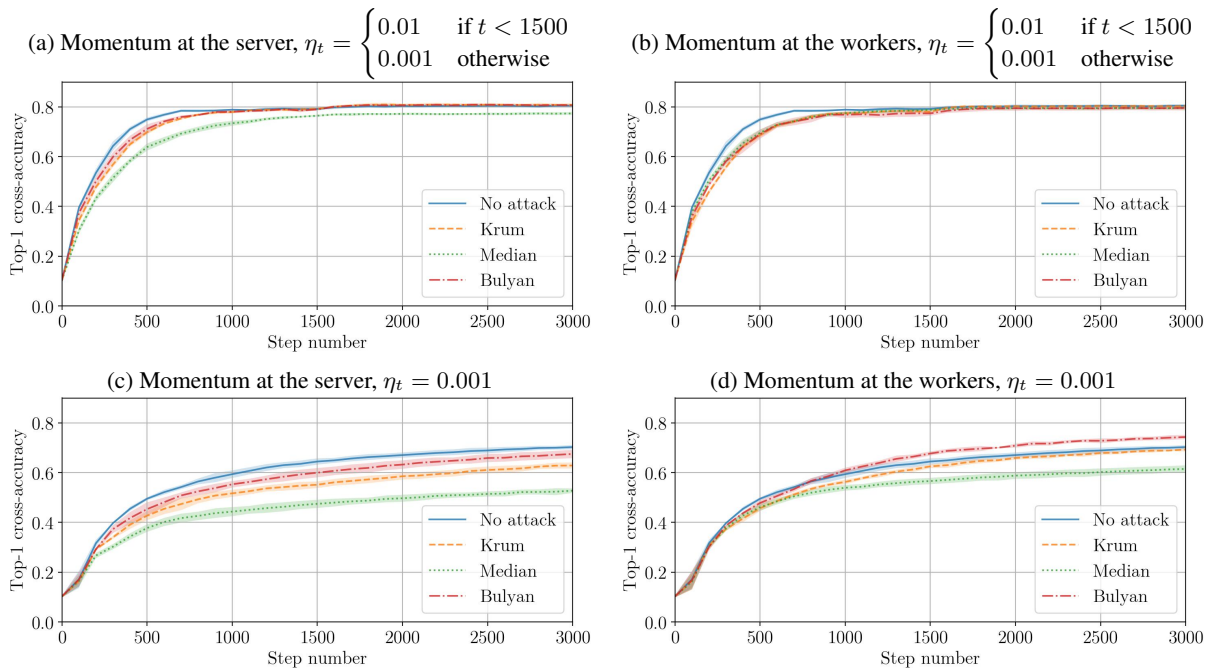

    \centering
    \begin{subfigure}[t]{0.45\linewidth}
        \centering
        \multicaption{Momentum at the server, \lrtitle{}}
		\includeplot{\detokenize{cifar10-empire-f_5-lr_0.01-at_update}}
	\end{subfigure}
	\quad
	\begin{subfigure}[t]{0.45\linewidth}
		\centering
		\multicaption{Momentum at the workers, \lrtitle{}}
		\includeplot{\detokenize{cifar10-empire-f_5-lr_0.01-at_worker}}
	\end{subfigure}
	\begin{subfigure}[t]{0.45\linewidth}
        \centering
        \multicaption{Momentum at the server, $\lr{t} = 0.001$}
		\includeplot{\detokenize{cifar10-empire-f_5-lr_0.001-at_update}}
	\end{subfigure}
	\quad
	\begin{subfigure}[t]{0.45\linewidth}
		\centering
		\multicaption{Momentum at the workers, $\lr{t} = 0.001$}
		\includeplot{\detokenize{cifar10-empire-f_5-lr_0.001-at_worker}}
	\end{subfigure}
    \caption{CIFAR-10 using $n = 25$ workers, including $f = 5$ Byzantine workers implementing \cite{empire}.
    This is the maximum number of Byzantine workers $\bulyan{}$ can support.
    With a reduced fraction of Byzantine workers to a quarter compared to Figure \ref{fig:xacc-cifar-empire-half}, the effect of the attack is almost void.
    Notably in this setting, \bulyan{} improves the cross-accuracy gain per step over \krum{} and \medianoid{}.}
    \label{fig:xacc-cifar-empire-quarter}
    \label{fig:xacc-last}
\end{figure*}

The results regarding the cross-accuracy are layed out by ``blocks'' of 4 experiment setups presenting the same model, dataset, number of Byzantine workers and attack.
These results are presented from figures \ref{fig:xacc-first} to \ref{fig:xacc-last}.
In each ``block'', the 2 top experiments use the larger learning rate and the 2 bottom ones the smaller, so looking below correspond to looking to the same experiment but with a smaller learning rate (and vice versa).
Similarly, the 2 left experiments use momentum at the server, while the 2 right ones use momentum at the workers, which allows for handy comparison of the effect of using one technique over the other.

\begin{figure*}
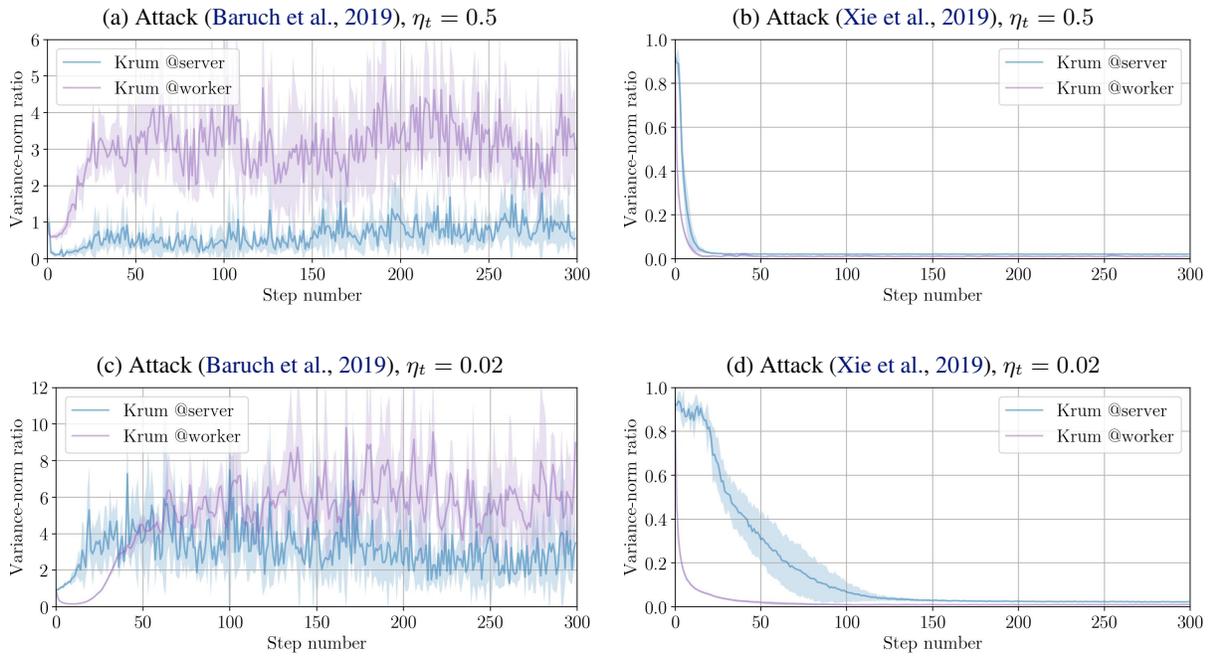

    \centering
    \begin{subfigure}[t]{0.45\linewidth}
        \centering
        \multicaption{Attack \cite{little}, $\lr{t} = 0.5$}
		\includeplot{\detokenize{mnist-little-krum-f_24-lr_0.5-ratio}}
	\end{subfigure}
	\quad
	\begin{subfigure}[t]{0.45\linewidth}
		\centering
		\multicaption{Attack \cite{empire}, $\lr{t} = 0.5$}
		\includeplot{\detokenize{mnist-empire-krum-f_24-lr_0.5-ratio}}
		\label{fig:ratio-mnist-krum-empire-large-half}
	\end{subfigure}
	\begin{subfigure}[t]{0.45\linewidth}
        \centering
        \multicaption{Attack \cite{little}, $\lr{t} = 0.02$}
		\includeplot{\detokenize{mnist-little-krum-f_24-lr_0.02-ratio}}
	\end{subfigure}
	\quad
	\begin{subfigure}[t]{0.45\linewidth}
		\centering
		\multicaption{Attack \cite{empire}, $\lr{t} = 0.02$}
		\includeplot{\detokenize{mnist-empire-krum-f_24-lr_0.02-ratio}}
		\label{fig:ratio-mnist-krum-empire-small-half}
	\end{subfigure}
    \caption{MNIST using $n = 51$ workers, including $f = 24$ Byzantine workers defended against by \krum{}.
    This setting contains the full range of behaviors one can observe in the subsequent figures.
    One first, notable behavior was predicted by the theory: until convergence is reached, reducing the learning rate decreases the variance-norm ratio.
    The second, recurrent behavior is that, when the model is driven useless (Figure \ref{fig:xacc-mnist-empire-half}), the ratio reaches low values.
    Such decreasing curves (\ref{fig:ratio-mnist-krum-empire-large-half} and \ref{fig:ratio-mnist-krum-empire-small-half}) are empirical, distinctive signal of a very successful attack.
    Indeed for a successful \emph{defense}, one should expect the ratio to grow to infinity, as the norm of the honest gradient goes toward 0.}
    \label{fig:ratio-first}
\end{figure*}

The results regarding the variance-norm ratio are also layed out by ``blocks'' of 4 experiment setups presenting the same model, dataset, number of Byzantine workers and defense.
These results are presented from figures \ref{fig:ratio-first} to \ref{fig:ratio-last}.
In each ``block'', the attack from \cite{little} is use on the left column and \cite{empire} on the right.
As for the cross-accuracy, the top row use the larger learning rate and the bottom row shows the effect of using a smaller one.

For figures \ref{fig:xacc-first} to \ref{fig:ratio-last}, the captions present to the reader the hyperparameters used in each of the experiments, along with comments about the observed behaviors.

\begin{figure*}
    \centering
    \begin{subfigure}[t]{0.45\linewidth}
        \centering
        \multicaption{Attack \cite{little}, $\lr{t} = 0.5$}
		\includeplot{\detokenize{mnist-little-krum-f_12-lr_0.5-ratio}}
	\end{subfigure}
	\quad
	\begin{subfigure}[t]{0.45\linewidth}
		\centering
		\multicaption{Attack \cite{empire}, $\lr{t} = 0.5$}
		\includeplot{\detokenize{mnist-empire-krum-f_12-lr_0.5-ratio}}
	\end{subfigure}
	\begin{subfigure}[t]{0.45\linewidth}
        \centering
        \multicaption{Attack \cite{little}, $\lr{t} = 0.02$}
		\includeplot{\detokenize{mnist-little-krum-f_12-lr_0.02-ratio}}
	\end{subfigure}
	\quad
	\begin{subfigure}[t]{0.45\linewidth}
		\centering
		\multicaption{Attack \cite{empire}, $\lr{t} = 0.02$}
		\includeplot{\detokenize{mnist-empire-krum-f_12-lr_0.02-ratio}}
	\end{subfigure}
    \caption{MNIST using $n = 51$ workers, including $f = 12$ Byzantine workers defended against by \krum{}.
    See Figure \ref{fig:ratio-first}.}
\end{figure*}

\begin{figure*}
    \centering
    \begin{subfigure}[t]{0.45\linewidth}
        \centering
        \multicaption{Attack \cite{little}, \lrtitle{}}
		\includeplot{\detokenize{cifar10-little-krum-f_11-lr_0.01-ratio}}
	\end{subfigure}
	\quad
	\begin{subfigure}[t]{0.45\linewidth}
		\centering
		\multicaption{Attack \cite{empire}, \lrtitle{}}
		\includeplot{\detokenize{cifar10-empire-krum-f_11-lr_0.01-ratio}}
	\end{subfigure}
	\begin{subfigure}[t]{0.45\linewidth}
        \centering
        \multicaption{Attack \cite{little}, $\lr{t} = 0.001$}
		\includeplot{\detokenize{cifar10-little-krum-f_11-lr_0.001-ratio}}
	\end{subfigure}
	\quad
	\begin{subfigure}[t]{0.45\linewidth}
		\centering
		\multicaption{Attack \cite{empire}, $\lr{t} = 0.001$}
		\includeplot{\detokenize{cifar10-empire-krum-f_11-lr_0.001-ratio}}
	\end{subfigure}
    \caption{CIFAR-10 using $n = 25$ workers, including $f = 11$ Byzantine workers defended against by \krum{}.
    See Figure \ref{fig:ratio-first}.}
\end{figure*}

\begin{figure*}
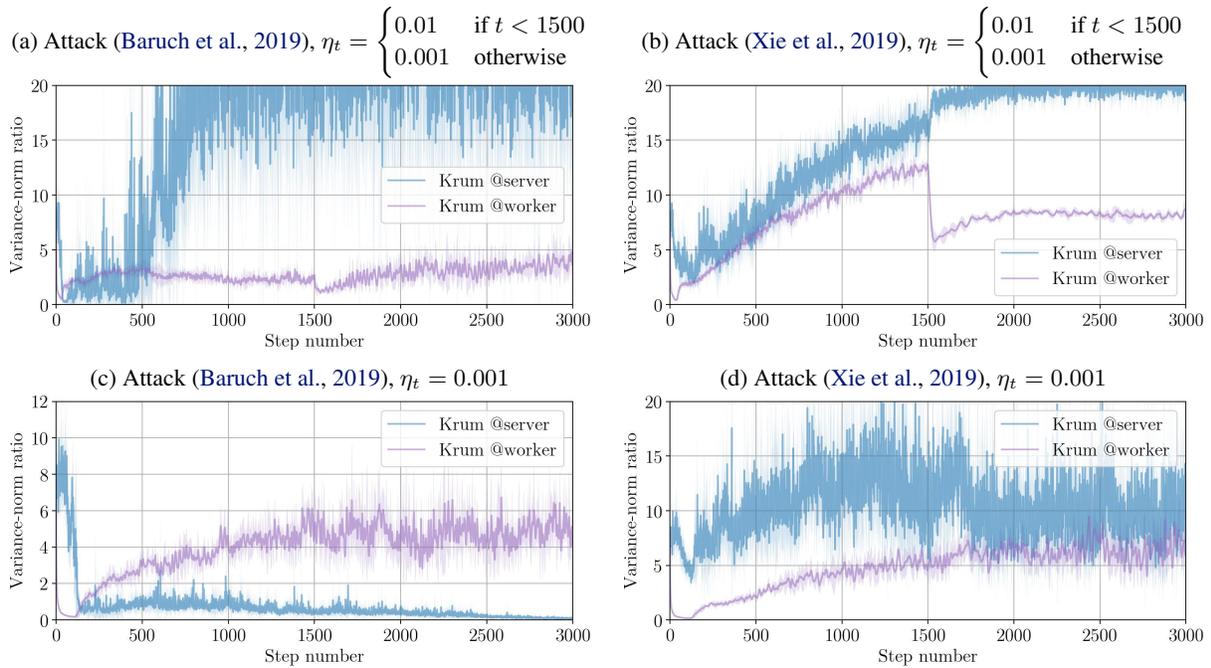

    \centering
    \begin{subfigure}[t]{0.45\linewidth}
        \centering
        \multicaption{Attack \cite{little}, \lrtitle{}}
		\includeplot{\detokenize{cifar10-little-krum-f_5-lr_0.01-ratio}}
	\end{subfigure}
	\quad
	\begin{subfigure}[t]{0.45\linewidth}
		\centering
		\multicaption{Attack \cite{empire}, \lrtitle{}}
		\includeplot{\detokenize{cifar10-empire-krum-f_5-lr_0.01-ratio}}
	\end{subfigure}
	\begin{subfigure}[t]{0.45\linewidth}
        \centering
        \multicaption{Attack \cite{little}, $\lr{t} = 0.001$}
		\includeplot{\detokenize{cifar10-little-krum-f_5-lr_0.001-ratio}}
	\end{subfigure}
	\quad
	\begin{subfigure}[t]{0.45\linewidth}
		\centering
		\multicaption{Attack \cite{empire}, $\lr{t} = 0.001$}
		\includeplot{\detokenize{cifar10-empire-krum-f_5-lr_0.001-ratio}}
	\end{subfigure}
    \caption{CIFAR-10 using $n = 25$ workers, including $f = 5$ Byzantine workers defended against by \krum{}.
    The ``fracture'' is due to the fact that the learning rate is decreased at step 1500.
    See Figure \ref{fig:ratio-first}.}
\end{figure*}

\begin{figure*}
    \centering
    \begin{subfigure}[t]{0.45\linewidth}
        \centering
        \multicaption{Attack \cite{little}, $\lr{t} = 0.5$}
		\includeplot{\detokenize{mnist-little-median-f_24-lr_0.5-ratio}}
	\end{subfigure}
	\quad
	\begin{subfigure}[t]{0.45\linewidth}
		\centering
		\multicaption{Attack \cite{empire}, $\lr{t} = 0.5$}
		\includeplot{\detokenize{mnist-empire-median-f_24-lr_0.5-ratio}}
	\end{subfigure}
	\begin{subfigure}[t]{0.45\linewidth}
        \centering
        \multicaption{Attack \cite{little}, $\lr{t} = 0.02$}
		\includeplot{\detokenize{mnist-little-median-f_24-lr_0.02-ratio}}
	\end{subfigure}
	\quad
	\begin{subfigure}[t]{0.45\linewidth}
		\centering
		\multicaption{Attack \cite{empire}, $\lr{t} = 0.02$}
		\includeplot{\detokenize{mnist-empire-median-f_24-lr_0.02-ratio}}
	\end{subfigure}
    \caption{MNIST using $n = 51$ workers, including $f = 24$ Byzantine workers defended against by \medianoid{}.
    See Figure \ref{fig:ratio-first}.}
\end{figure*}

\begin{figure*}
    \centering
    \begin{subfigure}[t]{0.45\linewidth}
        \centering
        \multicaption{Attack \cite{little}, $\lr{t} = 0.5$}
		\includeplot{\detokenize{mnist-little-median-f_12-lr_0.5-ratio}}
	\end{subfigure}
	\quad
	\begin{subfigure}[t]{0.45\linewidth}
		\centering
		\multicaption{Attack \cite{empire}, $\lr{t} = 0.5$}
		\includeplot{\detokenize{mnist-empire-median-f_12-lr_0.5-ratio}}
	\end{subfigure}
	\begin{subfigure}[t]{0.45\linewidth}
        \centering
        \multicaption{Attack \cite{little}, $\lr{t} = 0.02$}
		\includeplot{\detokenize{mnist-little-median-f_12-lr_0.02-ratio}}
	\end{subfigure}
	\quad
	\begin{subfigure}[t]{0.45\linewidth}
		\centering
		\multicaption{Attack \cite{empire}, $\lr{t} = 0.02$}
		\includeplot{\detokenize{mnist-empire-median-f_12-lr_0.02-ratio}}
	\end{subfigure}
    \caption{MNIST using $n = 51$ workers, including $f = 12$ Byzantine workers defended against by \medianoid{}.
    See Figure \ref{fig:ratio-first}.}
\end{figure*}

\begin{figure*}
    \centering
    \begin{subfigure}[t]{0.45\linewidth}
        \centering
        \multicaption{Attack \cite{little}, \lrtitle{}}
		\includeplot{\detokenize{cifar10-little-median-f_11-lr_0.01-ratio}}
	\end{subfigure}
	\quad
	\begin{subfigure}[t]{0.45\linewidth}
		\centering
		\multicaption{Attack \cite{empire}, \lrtitle{}}
		\includeplot{\detokenize{cifar10-empire-median-f_11-lr_0.01-ratio}}
	\end{subfigure}
	\begin{subfigure}[t]{0.45\linewidth}
        \centering
        \multicaption{Attack \cite{little}, $\lr{t} = 0.001$}
		\includeplot{\detokenize{cifar10-little-median-f_11-lr_0.001-ratio}}
	\end{subfigure}
	\quad
	\begin{subfigure}[t]{0.45\linewidth}
		\centering
		\multicaption{Attack \cite{empire}, $\lr{t} = 0.001$}
		\includeplot{\detokenize{cifar10-empire-median-f_11-lr_0.001-ratio}}
	\end{subfigure}
    \caption{CIFAR-10 using $n = 25$ workers, including $f = 11$ Byzantine workers defended against by \medianoid{}.
    See Figure \ref{fig:ratio-first}.}
\end{figure*}

\begin{figure*}
    \centering
    \begin{subfigure}[t]{0.45\linewidth}
        \centering
        \multicaption{Attack \cite{little}, \lrtitle{}}
		\includeplot{\detokenize{cifar10-little-median-f_5-lr_0.01-ratio}}
	\end{subfigure}
	\quad
	\begin{subfigure}[t]{0.45\linewidth}
		\centering
		\multicaption{Attack \cite{empire}, \lrtitle{}}
		\includeplot{\detokenize{cifar10-empire-median-f_5-lr_0.01-ratio}}
	\end{subfigure}
	\begin{subfigure}[t]{0.45\linewidth}
        \centering
        \multicaption{Attack \cite{little}, $\lr{t} = 0.001$}
		\includeplot{\detokenize{cifar10-little-median-f_5-lr_0.001-ratio}}
	\end{subfigure}
	\quad
	\begin{subfigure}[t]{0.45\linewidth}
		\centering
		\multicaption{Attack \cite{empire}, $\lr{t} = 0.001$}
		\includeplot{\detokenize{cifar10-empire-median-f_5-lr_0.001-ratio}}
	\end{subfigure}
    \caption{CIFAR-10 using $n = 25$ workers, including $f = 5$ Byzantine workers defended against by \medianoid{}.
    See Figure \ref{fig:ratio-first}.}
\end{figure*}

\begin{figure*}
    \centering
    \begin{subfigure}[t]{0.45\linewidth}
        \centering
        \multicaption{Attack \cite{little}, $\lr{t} = 0.5$}
		\includeplot{\detokenize{mnist-little-bulyan-f_12-lr_0.5-ratio}}
	\end{subfigure}
	\quad
	\begin{subfigure}[t]{0.45\linewidth}
		\centering
		\multicaption{Attack \cite{empire}, $\lr{t} = 0.5$}
		\includeplot{\detokenize{mnist-empire-bulyan-f_12-lr_0.5-ratio}}
	\end{subfigure}
	\begin{subfigure}[t]{0.45\linewidth}
        \centering
        \multicaption{Attack \cite{little}, $\lr{t} = 0.02$}
		\includeplot{\detokenize{mnist-little-bulyan-f_12-lr_0.02-ratio}}
	\end{subfigure}
	\quad
	\begin{subfigure}[t]{0.45\linewidth}
		\centering
		\multicaption{Attack \cite{empire}, $\lr{t} = 0.02$}
		\includeplot{\detokenize{mnist-empire-bulyan-f_12-lr_0.02-ratio}}
	\end{subfigure}
    \caption{MNIST using $n = 51$ workers, including $f = 12$ Byzantine workers defended against by \bulyan{}.
    See Figure \ref{fig:ratio-first}.}
\end{figure*}

\begin{figure*}
    \centering
    \begin{subfigure}[t]{0.45\linewidth}
        \centering
        \multicaption{Attack \cite{little}, \lrtitle{}}
		\includeplot{\detokenize{cifar10-little-bulyan-f_5-lr_0.01-ratio}}
	\end{subfigure}
	\quad
	\begin{subfigure}[t]{0.45\linewidth}
		\centering
		\multicaption{Attack \cite{empire}, \lrtitle{}}
		\includeplot{\detokenize{cifar10-empire-bulyan-f_5-lr_0.01-ratio}}
	\end{subfigure}
	\begin{subfigure}[t]{0.45\linewidth}
        \centering
        \multicaption{Attack \cite{little}, $\lr{t} = 0.001$}
		\includeplot{\detokenize{cifar10-little-bulyan-f_5-lr_0.001-ratio}}
	\end{subfigure}
	\quad
	\begin{subfigure}[t]{0.45\linewidth}
		\centering
		\multicaption{Attack \cite{empire}, $\lr{t} = 0.001$}
		\includeplot{\detokenize{cifar10-empire-bulyan-f_5-lr_0.001-ratio}}
	\end{subfigure}
    \caption{CIFAR-10 using $n = 25$ workers, including $f = 5$ Byzantine workers defended against by \bulyan{}.
    See Figure \ref{fig:ratio-first}.}
    \label{fig:ratio-last}
\end{figure*}

\end{document}